\documentclass[10pt,journal,compsoc]{IEEEtran}
\usepackage{microtype}
\usepackage{graphicx}
\usepackage{subfigure}
\usepackage{booktabs} 

\usepackage{times}
\usepackage{epsfig}
\usepackage{amsmath}
\usepackage{amssymb}
\usepackage{color}
\usepackage[T1]{fontenc}
\usepackage{multirow}
\usepackage{algorithm,algpseudocode,amsmath}
\usepackage{verbatim}

\graphicspath{{./Images/}}

\ifCLASSOPTIONcompsoc
  \usepackage[nocompress]{cite}
\else
  \usepackage{cite}
\fi
\ifCLASSINFOpdf
\else
\fi
\begin{document}
%
\title{Sill-Net: Feature Augmentation with Separated Illumination Representation}
%
%
%
%

\author{Haipeng~Zhang,~\IEEEmembership{Student Member,~IEEE,}
        Zhong~Cao,~\IEEEmembership{Student Member,~IEEE,}
        Ziang~Yan,~\IEEEmembership{}
        and~Changshui~Zhang,~\IEEEmembership{Fellow,~IEEE}
\IEEEcompsocitemizethanks{\IEEEcompsocthanksitem H. Zhang, Z. Cao, Z. Yan and C. Zhang are with the Institute for Artificial Intelligence, Tsinghua University (THUAI), Beijing 100084, China, and also with the State Key Laboratory of Intelligence Technologies and Systems, Beijing National Research Center for Information Science and Technologies (BNRist) , Department of Automation, Tsinghua University, Beijing 100084, China (e-mail: zhanghp16@mails.tsinghua.edu.cn; caozhong14@mails.tsinghua.edu.cn; yza18@mails.tsinghua.edu.cn; zcs@mail.tsinghua.edu.cn). \protect\\
\IEEEcompsocthanksitem Corresponding Author: H. Zhang and C. Zhang.}
\thanks{}}

%
%

\markboth{IEEE TRANSACTIONS ON PATTERN ANALYSIS AND MACHINE INTELLIGENCE}%
{Shell \MakeLowercase{\textit{et al.}}: Bare Demo of IEEEtran.cls for Computer Society Journals}
%



\IEEEtitleabstractindextext{%
\begin{abstract}
For visual object recognition tasks, the illumination variations can cause distinct changes in object appearance and thus confuse the deep neural network-based recognition models.
 Especially for some rare illumination conditions, collecting sufficient training samples could be time-consuming and expensive.
 To solve this problem, in this paper we propose a novel neural network architecture called \textbf{S}eparating-\textbf{Ill}umination \textbf{Net}work (Sill-Net).
 Sill-Net learns to separate illumination features from images, and then we augment training samples with these separated illumination features in the feature space.
 The model is further trained on the augmented samples to be robust to illumination variations.
 We provide a fresh perspective to focus on removing the semantic part of images and storing the illumination to the repository for augmentations instead of augmenting the semantic part of features.
 Extensive experimental results demonstrate that our approach outperforms current state-of-the-art methods on common object classification benchmarks.
\end{abstract}

\begin{IEEEkeywords}
Feature augmentation, illumination representation, one/few-shot learning, feature disentanglement, image reconstruction.
\end{IEEEkeywords}}

\maketitle

\IEEEdisplaynontitleabstractindextext

%
\IEEEpeerreviewmaketitle

\IEEEraisesectionheading{\section{Introduction}\label{sec:introduction}}

%
%
%
%
\IEEEPARstart{A}{lthough} deep learning models have achieved remarkable successes in various computer vision tasks \cite{krizhevsky2017imagenet, simonyan2014very, russakovsky2015imagenet, he2016deep}, vast amounts of annotated training data are usually required for superior performance in most scenarios.
For object classification tasks, the requirement for a large training set could be partially explained by the fact that many latent variables (e.g., positions/postures of the objects, the brightness/contrast of the image, and the illumination conditions) can cause significant changes in the appearance of objects.
Although collecting a large training set to cover all possible values of these latent variables could improve the recognition performance, for rare latent values such as extreme illumination conditions it could be prohibitively time-consuming and expensive to collect enough training images \cite{zhang2018improving}.

In this paper, we restrict our attention to illumination conditions.
For many real-world computer vision applications (e.g., autonomous driving and video surveillance) it is essential to recognize the objects under extreme illumination conditions such as backlighting, overexposure, and other complicated cast shadows.
Thus, we reckon it is desirable to improve recognition models' generalization ability under different illumination conditions in order to deploy robust models in real-world applications.

\begin{figure}[htbp]
\begin{center}
\vspace{-12mm}
\centerline{\includegraphics[width=\columnwidth]{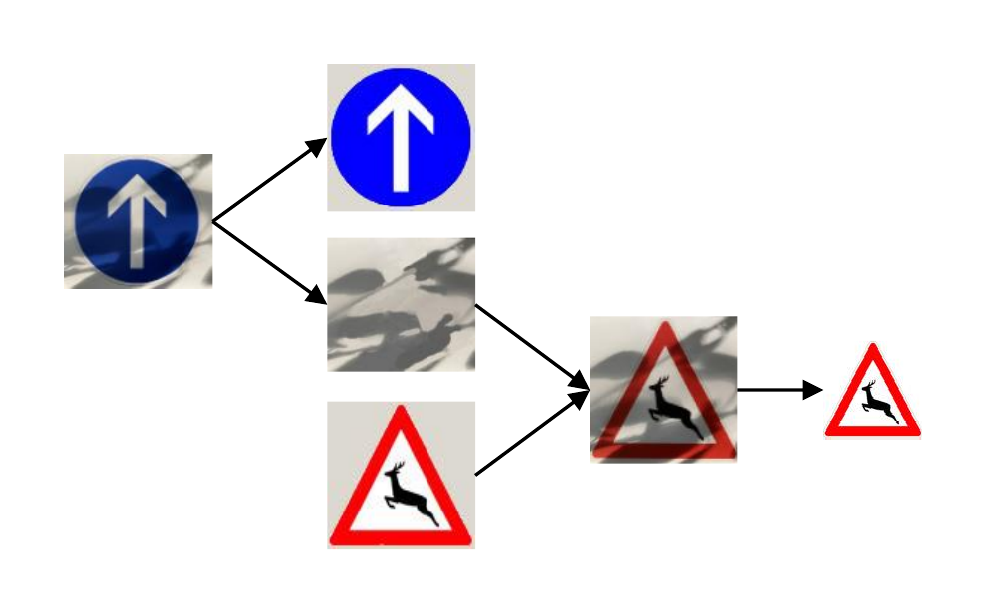}}
\vspace{-6mm}
\caption{Illustration of the key idea of our approach. The semantic and illumination representation are separated from the training image (mandatory straight). The illumination representation is used to augment the support sample (deer crossing).}
\label{fig:idea}
\end{center}
\vspace{-4mm}
\end{figure}

We propose a novel neural network architecture named \textit{\textbf{S}eparating-\textbf{Ill}umination \textbf{Net}work (Sill-Net)} to deal with such problems.
The key idea of our approach is to separate the illumination features from the semantic features in images, and then augment the separated illumination features onto other training samples (hereinafter we name these samples as ``support samples'') to construct a more extensive feature set for subsequent training (see Figure \ref{fig:idea}).
Specifically, our approach consists of three steps.
In the first step, we separate the illumination and semantic features for all images in the existing dataset via a disentanglement method, and use the separated illumination features to build an illumination repository.
Then, we transplant the illumination repository to the support samples to construct an augmented training set and use it to train a recognition model.
Finally, test images are fed into the trained model for classification.
Our proposed approach could improve the robustness to illumination conditions since the support samples used for training are blended with many different illumination features.
Thus after training, the obtained model would naturally generalize better under various illumination conditions.

To separate better illumination features, we train the model on datasets with simple symbolic objects and diverse illumination in the separation step. The separated illumination features can be used for augmentation on extensive datasets including symbolic objects and general image datasets.
Our method is evaluated in both traditional and one/few-shot classification tasks.


\begin{figure*}[htbp]
 \begin{center}
 \vspace{-5mm}
  \includegraphics[width=0.9\textwidth]{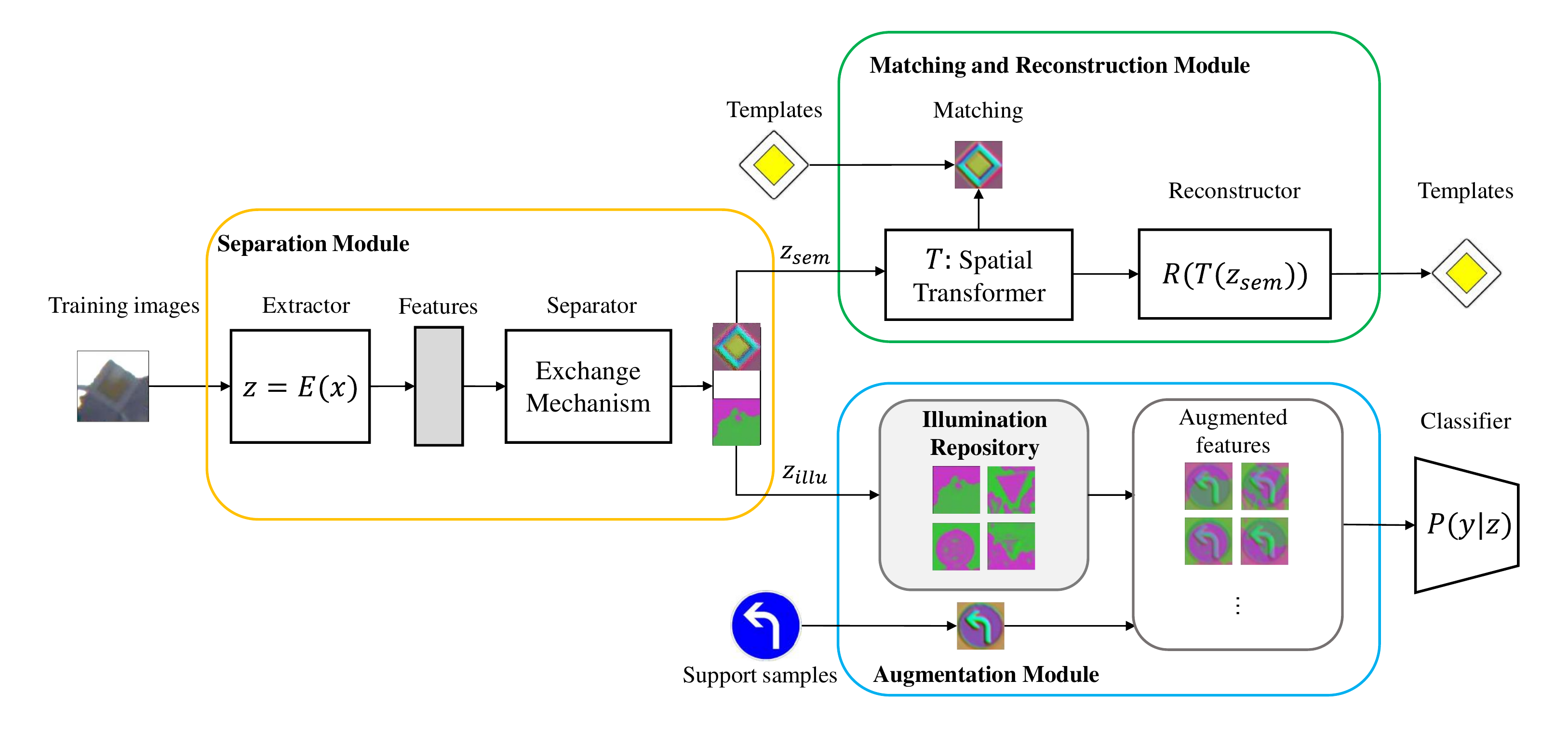}
 \end{center}
 \vspace{-6mm}
 \caption{Illustration of the architecture of Sill-Net. Sill-Net consists of three main modules: the separation module, the matching and reconstruction module, and the augmentation module. The semantic and illumination features are separated by the exchange mechanism in the first module. The semantic features are constrained to be informative by the matching and reconstruction module. The illumination features are stored into a repository. In the augmentation module, we use the illumination features in the repository to augment the support samples (e.g., template images) for training a generalizable model.}
 \label{fig:model}
 \vspace{-4mm}
\end{figure*}

Our contributions are summarized as follows:

1) We develop a novel algorithm to separate the semantic and illumination features from images. The separated illumination features construct an illumination feature repository, which could effortlessly enhance the illumination variety of training data and thus improving the robustness to illumination conditions of the trained deep model. The codes and repository are released \footnote{The codes are released at \textit{https://github.com/lanfenghuanyu/Sill-Net}.}.

2) We provide a fresh perspective to focus on removing the semantic part of images and storing the remaining (illumination) to the repository for augmentations while recent works concentrate on how to augment the semantic part of features or learn the policies.


3) We evaluate Sill-Net on several object classification benchmarks, i.e., traffic datasets (GTSRB, TT100K, CTSD, etc.), logo datasets (Belgalogos, FlickrLogos32, and TopLogo-10) and generalized one/few-shot benchmarks (\emph{mini}ImageNet, CUB, CIFAR-FS). Sill-Net outperforms the state-of-the-art (SOTA) methods by a large margin in most cases.

\vspace{-3mm}

\section{Proposed Method}\label{sec:method}

In this section, we introduce our \textit{\textbf{S}eparating-\textbf{Ill}umination \textbf{Net}work (Sill-Net)}. Sill-Net first learns to separate the semantic and illumination features of training images. Then the illumination features are blended with the semantic feature of each support sample to construct an augmented feature set. Finally, we train again on the illumination-augmented feature set for classification. 

The architecture of Sill-Net is illustrated in Figure \ref{fig:model}. It mainly consists of the following modules: the separation module, the matching and reconstruction module, and the augmentation module. In detail, we implement Sill-Net in three phases (see Algorithm \ref{algo:1}):

1) In the separation phase, the separation module is trained to separate the features into semantic parts and illumination parts for all training images. The matching and reconstruction module promotes the learning of better semantic feature representation.
The learned illumination features are stored into an illumination repository. The details are illustrated in Section \ref{sec:separation}.

2) In the augmentation phase, the semantic feature of each support image is combined with all illumination features in the repository to build an augmented feature set to train the classifier. The augmentation module is illustrated in Section \ref{sec:augmentation}.

3) During inference, test images are input into the well-trained model to be predicted in an end-to-end manner (see Section \ref{sec:test}).

This approach assumes that the illumination distribution learned from training data is able to characterize that of test data. Thus the illumination features can be used as feature augmentation for sufficient training.

We choose different support samples in different visual tasks. For instance, in traditional classification tasks, we use the training images themselves as support samples; in one-shot classification tasks, we construct the support set with template images (i.e., graphic symbols visually and abstractly representing semantic information) available from official websites or wikis.

\vspace{-3mm}
\subsection{Separate semantic and illumination features} \label{sec:separation}

Let $\mathcal{X}=\left\{(\mathbf{x}_i, y_i, \mathbf{t}_i)\right\}_{i=1}^{N}$ represent the labeled dataset of training classes with $N$ images, where $\mathbf{x}_i$ denotes the $i$-th training image, $y_i$ is the one-hot label, and  $\mathbf{t_i}$ denotes the corresponding template image (or any other image of the object without much deformation).

{\bf Separation module} A feature extractor denoted by $\mathbf{z}={E}(\mathbf{x})$ learns the separated features $\mathbf{z}$ from images  $\mathbf{x}$, where $\mathbf{z}$ is artifically split along channels: $\mathbf{z}\rightarrow[\mathbf{z}_{sem}, \mathbf{z}_{illu}]$. We expect $\mathbf{z}_{sem}$ as the semantic feature to learn the consistent information of the same category during training, while $\mathbf{z}_{illu}$ as the illumination feature to learn various illumination information.


We first specify what illumination represents in our paper. Illumination is one of the environmental impacts causing appearance changes but no label changes. It should be noted that illumination is always related to or influenced by objects both in the foreground and background. Here we call the features most relevant to general illumination impacts such as light and shadows, but not category-specific as illumination features. Technically, we divide the object feature into different channels, one half learning to determine the category label defined as the semantic feature, and the other half reflecting illumination variations but unrelated to the category label defined as the illumination feature. To ensure that the learned features contain minimal irrelevant information beyond labeled objects and illumination, we choose training datasets with various illumination conditions and few confounding objects.

To achieve the separation, we design our model according to the following three basic requirements:

1) Semantic features can predict labels while illumination features can not.

2) Semantic features are informative to reconstruct the corresponding template images.

3) Illumination features contain minimal semantic information.

\begin{figure}[t]
 \begin{center}
  \vspace{-4mm}
  \includegraphics[width=\columnwidth]{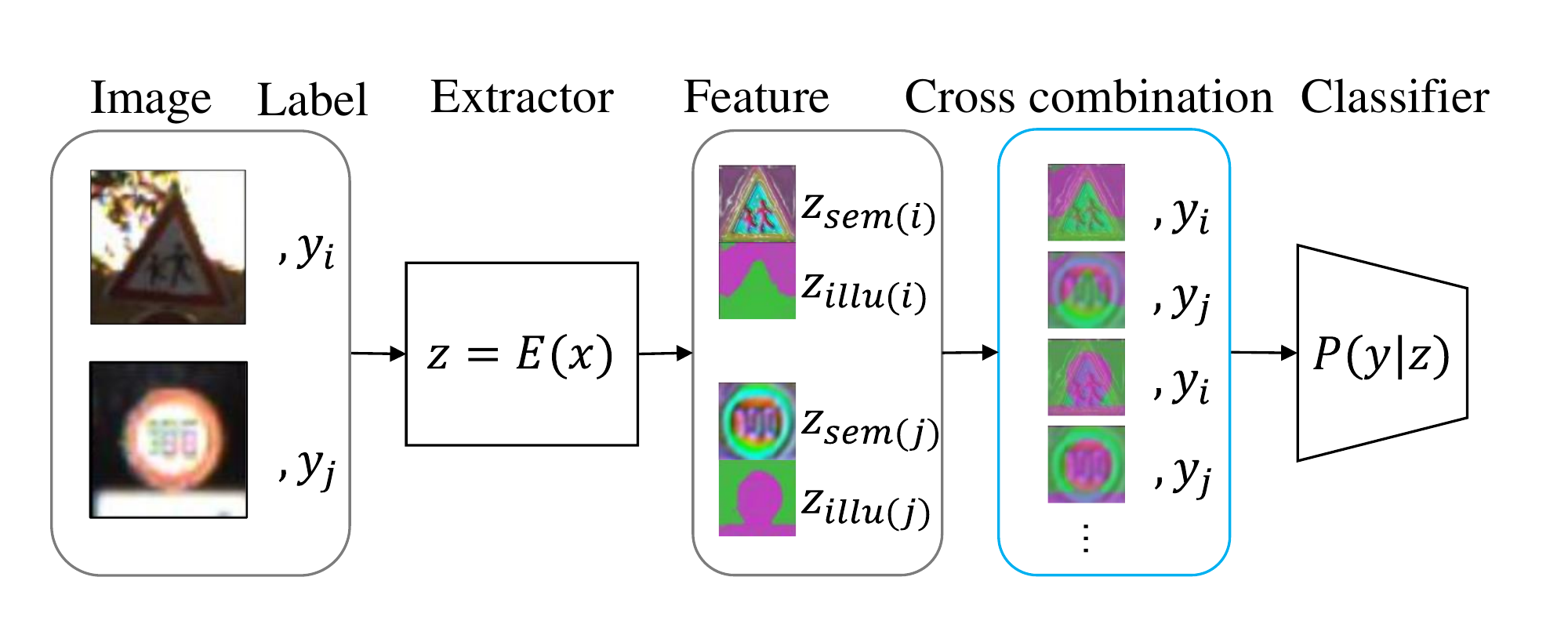}
 \end{center}
 \vspace{-6mm}
 \caption{Illustration of the exchange mechanism. The semantic and illumination features are exchanged between random paired images with labels $y_i$ and $y_j$. Then we obtain cross combined features labeled the same as the images corresponding to the semantic features. These features are then classified as the specified labels.}
 \label{fig:exchange}
 \vspace{-2mm}
\end{figure}

{\bf Exchange mechanism.} To fulfill the first requirement that semantic features can predict labels while illumination features can not, we utilize a feature exchange mechanism enlightened by \cite{xiao2018elegant} to separate the features. The semantic feature ${\mathbf{z}_{sem}}_{(i)}$ of one image $\mathbf{x}_i$ is blended with the illumination feature ${\mathbf{z}_{illu}}_{(j)}$ of another image $\mathbf{x}_j$ to form a new one through feature mixup \cite{zhang2017mixup} (see Figure \ref{fig:exchange}):

\vspace{-2mm}
\begin{equation}
 \mathbf z = r{{\mathbf z}_{sem}}_{(i)} + (1-r) {{\mathbf z}_{illu}}_{(j)},
\end{equation}

\noindent where the proportion $r\in (0,1)$. Generally, we set $r=0.5$.

As required, the blended feature $\mathbf{z}$ retains the same label $y_i$ as the semantic feature. 
It means that $\mathbf{z}$ should contain more information representing $y_i$ and less information representing $y_j$. 
Hence through subsequent training, the semantic information of $\mathbf{x}_i$ would be retained in the semantic feature ${\mathbf{z}_{sem}}_{(i)}$ while the semantic information of $\mathbf{x}_j$ would be reduced in the illumination feature ${\mathbf{z}_{illu}}_{(j)}$.

We implement the exchange process for random pairs of images, building a new exchanged feature set:

\vspace{-4mm}
\begin{equation}
 \mathcal{Z}_{exc}=\left\{ r{{\mathbf z}_{sem}}_{(i)} + (1-r) {{\mathbf z}_{illu}}_{(j)}, y_i \bigm| i,j=1,\cdots,N \right\}.
\end{equation}

The mixed features are then input into a classifier $P$ for label prediction. We denote the distribution of the predicted label $y$ given the mixed feature $\mathbf z$ by $P(y | \mathbf z)$. Then we minimize the cross-entropy loss:

\vspace{-6mm}
\begin{equation}
 \mathcal{L}_{exc}=-\frac{1}{N_{exc}} \sum_{i=1}^{N_{exc}} \sum_{c=1}^{M} y_{ic} \log P\left(y_{ic}|{\mathbf z}_i \in \mathcal{Z}_{exc} \right),
\end{equation}
\vspace{-3mm}

\noindent where $N_{exc}=|\mathcal{Z}_{exc}|$ denotes the number of recombined features in the augmented feature set, $M$ represents the class number of all images for training and test, and $y_{ic}$ is the $c$-th element of the one-hot label $y_i$.

After training by the exchange mechanism, the semantic information would be retained in semantic features but reduced in illumination features for all images.

{\bf Matching and reconstruction module.} To fulfill the second requirement, we first construct a matching module (as shown in Figure \ref{fig:model}) to make the semantic feature informative. Since we design the extractor without downsampling operations to maintain the spatial information of the object, the semantic feature of each training image should be similar to that of its corresponding template image. However, due to different camera distances and angles, real-world images are usually deformed in size and position compared to regular template images. 
Therefore, we adopt a spatial transformer $\mathcal{T}$ \cite{jaderberg2015spatial} to learn the parameters of affine transformation on the deformed objects and then rectify their semantic features to the regular positions (aligned with the templates).
We constrain the transformed semantic feature $\mathcal{T} \left( {{\mathbf z}_{sem}}_{(i)}|_{{\mathbf x}_i} \right)$ to be consistent with the template feature ${{\mathbf z}_{sem}}_{(i)}|_{{\mathbf t}_i}$ by the mean square error (MSE):

\vspace{-4mm}
\begin{equation}
 \mathcal{L}_{match}=\frac{1}{N} \sum_{i=1}^{N} \left( \mathcal{T} \left( {{\mathbf z}_{sem}}_{(i)}|_{{\mathbf x}_i} \right)-{{\mathbf z}_{sem}}_{(i)}|_{{\mathbf t}_i} \right )^2.
\end{equation}
\vspace{-3mm}

Then we design a reconstructor $\hat{\mathbf{t}}=R(\mathcal{T}({\mathbf{z}}_{sem}))$ (see Figure \ref{fig:model}) to retrieve the template image $\mathbf{t}$ from the semantic feature ${\mathbf{z}}_{sem}$ to ensure that it is informative enough. We constrain the reconstructed template ${\hat{\mathbf t}_i}$ by binary cross-entropy (BCE) loss:

\vspace{-4mm}
\begin{equation}
  \mathcal{L}_{recon}=\frac{1}{N} \sum_{i=1}^{N} \sum_{j} -t_{ij} \log \hat{t}_{ij} - \left(1-t_{ij}\right) \log \left(1-\hat{t}_{ij}\right),
\end{equation}
\vspace{-3mm}

\noindent where $t_{ij}$ represents the $j$-th pixel of the $i$-th template image ${\mathbf t}_i$. Since the template images are composed of primary colors within the range of $[0,1]$, binary cross-entropy (BCE) loss is sufficiently efficient for the retrieval \cite{kim2019variational}. So far, the semantic feature is constrained to be consistent with the template feature and informative enough to be reconstructed to its template image.

{\bf Constraints on illumination features.} According to the third requirement, it is essential to impose additional constraints on illumination features to reduce the semantic information. However, it is difficult to find suitable restrictions since the generally used datasets have no illumination labels.

Enlightened by the disentanglement metric proposed in \cite{suter2019robustly}, we design a constraint on illumination features by negative Post Interventional Disagreement (PIDA). Given a subset $\mathcal{X}_c=\left\{(\mathbf{x}_{ci}, y_{c})\right\}_{i=1}^{N_c}$ including $N_c$ images of the same label $y_c$, we write the loss as follows:

\vspace{-4mm}

\begin{equation}
 \begin{aligned}
  \mathcal{L}_{illu} & = -PIDA \\ &= -\sum_{c=1}^{M} \sum_{i=1}^{{N}_c}\mathcal{D} \left( \mathbb{E} \left({\mathbf z}_{illu}|_{\mathbf{x}_c, y_c} \right),{\mathbf z}_{illu}|_{{\mathbf x}_{ci}, y_c} \right),
 \end{aligned}
 \label{eq:pida}
\end{equation}

\noindent here, $\mathcal{D}$ is a proper distance function (e.g., $\ell_{2}$ -norm), ${\mathbf z}_{illu}|_{{\mathbf x}_{ci}, y_{c}}$ is the illumination feature of image ${\mathbf x}_{ci}$ with the same label $y_c$, $\mathbb{E}$ is the expectation, and ${N}_c$ is the number of images in class $c$.

According to Eq.\ (\ref{eq:pida}), PIDA quantifies the distances between the illumination feature of each same-labeled image ${\mathbf z}_{illu}|_{{\mathbf x}_{ci}, y_{c}}$ and their expectation $\mathbb{E} \left({\mathbf z}_{illu}|_{{\mathbf x}_c, y_c}\right)$ when the illumination conditions are changed. In the subset $\mathcal{X}_c$, the semantic information of each image is similar while the illumination information is different. Suppose an undesirable situation that the illumination features capture much semantic information rather than illumination information. The expectation would strengthen the common semantic component and weaken the distinct illumination components, and thus PIDA would be small. It means that the smaller PIDA is, the more semantic information the illumination feature captures compared to illumination information. By maximizing PIDA (i.e., minimizing ${L}_{illu}$), we can effectively reduce the common semantic information remaining in the illumination features. Besides, the constraint also avoids a trivial solution (all zeros) for the illumination features, since otherwise PIDA would be zero.

In summary, the overall loss function in the separation phase can be written as:

\vspace{-3mm}
\begin{equation}
 \mathcal{L}= \mathcal{L}_{exc} + \mathcal{L}_{match} + \mathcal{L}_{recon} + \mathcal{L}_{illu}.
\end{equation}

\noindent Through the above training, the model learns to separate the features into semantic and illumination features.

\vspace{-3mm}
\subsection{Augment samples by the illumination repository} \label{sec:augmentation}

After the separation phase, the illumination features are separated from training images. These features are collected to construct an illumination repository, expressed as follows:

\begin{equation}
 \mathcal{Z}_{illu}=\left\{ {\mathbf{z}_{illu}}_{(i)} \right\}_{i=1}^{N}.
\end{equation}

In the augmentation phase, we then use the illumination features to augment the support samples by a multiple of the repository size $N$. Consider $\mathcal{X}^{tst}=\left\{(\mathbf{x}_i^{tst}, y_i^{tst}, \mathbf{t}_i^{tst})\right\}_{i=1}^{N^{tst}}$ with $N^{tst}$ images of label $y^{tst}$, here we assume that the template images $\mathbf{t}_i^{tst}$ constitute the support set. We combine all illumination features in the repository with the semantic feature of each template ${\mathbf{z}_{sem}^{sup}}_{(i)} = E(\mathbf{t}_i^{tst})$ by feature mixup, building an augmented feature set as follows:

\vspace{-3mm}
\begin{small}
 \begin{equation}
  \mathcal{Z}_{aug} = \left\{ r {\mathbf{z}_{sem}^{sup}}_{(i)} + (1-r) {{\mathbf z}_{illu}}_{(j)}, y_i^{tst} \bigm| i=1,\cdots,N^{tst} \right\},
 \end{equation}
\end{small}
\vspace{-4mm}

\noindent where ${{\mathbf z}_{illu}}_{(j)} \in \mathcal{Z}_{illu}$ and $r=0.5$ by default.

We train the model again on the feature set $\mathcal{Z}_{aug}$. Despite the limited support samples provided, the model can be trained on the augmented feature set blended with various illumination features, making it generalizable to test data. The classification loss of augmented training is expressed as follows:

\vspace{-4mm}
\begin{equation}
 \mathcal{L}_{aug}=-\frac{1}{N_{aug}} \sum_{i=1}^{N} \sum_{c=1}^{M} y_{ic} \log P\left(y_{ic}|{\mathbf z}_i \in \mathcal{Z}_{aug} \right),
\end{equation}

\noindent where $N_{aug}=|\mathcal{Z}_{aug}|$ denotes the number of all recombined features in the augmented feature set.

Now, the model has been trained to be generalizable for testing.

\vspace{-3mm}
\subsection{Inference} \label{sec:test}

The feature extractor and classifier have been fully trained after the first two phases. Given the $i$-th test image $\mathbf{x}_i^{tst}$, the separated feature is extracted as ${\mathbf{z}}_{i} = E(\mathbf{x}_i^{tst})$. Then the classifier outputs the category label $\hat{c}$, formulated as:

\vspace{-2mm}
\begin{equation}
 \hat{c}_i = \underset{c} {\arg \max } \  P \left(y_{ic}|{\mathbf z}_i \right).
\end{equation}

\noindent It should be noted that we use the mixed features instead of the semantic ones for inference, which is consistent with the training of the classifier during the augmentation phase.
The inference is achieved in an end-to-end manner.

Overall, the main pipeline of Sill-Net is shown in Algorithm \ref{algo:1}. Due to the page limit, detailed architectures and parameter settings of the network are described in Appendix A.1.

\vspace{-2mm}


\subsection{Manipulations on the illumination repository} \label{sec:manipulation}

We further manipulate the illumination repository to improve the performance and interpretability of our method.

\subsubsection{Increase the diversity and reduce the size}\label{sec:repository}
Two factors affect the quality of the illumination repository: the diversity of illumination characteristics and the size.

We previously assumed that our separated illumination repository is able to characterize the illumination distribution of test data. To better fulfill this assumption, we have to further increase the diversity of the basic repository.
Assumed that all the illumination features are from a feature space, so we linearly combine these illumination features to generate new ones. For simplicity, we linearly interpolate random pairs of illumination features to form an expanded repository:



\vspace{-4mm}
\begin{equation}
\begin{aligned}
\mathcal{Z}_{i l l u}^{\exp }=\{& \gamma \mathbf{z}_{i l l u(i)}+(1-\gamma) \mathbf{z}_{i l l u(j)} \mid \\
&\left.\mathbf{z}_{i l l u(i)}, \mathbf{z}_{i l l u(j)} \in \mathcal{Z}_{i l l u}, \gamma \in[0,1]\right\},
\end{aligned}
\end{equation}

\noindent where $\gamma$ is randomly sampled from a uniform distribution. Theoretically, we can expand the repository to any size $N_{exp}$.

Furthermore, the size of the illumination repository directly affects the time cost of the application. When the illumination repository is very large, the calculation in the augmentation phase is time-consuming. 
To increase the training speed, we try to compress the illumination repository without decreasing much of the accuracy of image recognition.
As observed, some training images are obtained under similar illumination conditions, so that their illumination features will be very similar. We can choose the representative illumination features to replace the similar ones, so that the illumination repository can be compressed. Clustering is a simple and useful method to find representative repository features avoiding much information loss caused by data selection.

Now we have constructed an illumination repository with representative and diverse illumination features via selection and expansion. Our illumination repository is pre-stored and agnostic to the applied datasets of new tasks. 
Assume that the manipulated repository basically characterizes the illumination distribution, feature augmentation with these illumination features can facilitate not only symbolic objects recognition but also general few-shot classification. 
See Section \ref{sec:expansion} and \ref{sec:few-shot} for experimental details.


\subsubsection{Illumination reconstruction and transplantation}

So far, all our manipulations have been performed in feature space. To further interpret the separated illumination features, we reconstruct them back to visual images.

Previously, we have separated the semantic and illumination features of training images $\mathbf{x}$. Then we train another reconstructor to retrieve real-world images from the separated features, expressed as $\hat{\mathbf{x}}=R_{real}({\mathbf{z}}_{sem},{\mathbf{z}}_{illu})$. We adopt the binary cross-entropy (BCE) as the reconstruction loss the same as before: 

\vspace{-4mm}
\begin{equation}
 \mathcal{L}_{recon}^{real}=\frac{1}{N} \sum_{i=1}^{N} \sum_{j} -x_{ij} \log \hat{x}_{ij} - \left(1-x_{ij}\right) \log \left(1-\hat{x}_{ij}\right),
\end{equation}

\noindent where $x_{ij}$ represents the $j$-th pixel of the $i$-th training image ${\mathbf x}_i$.

After training, we can transplant certain illumination features to any support samples and generate new images with realistic illumination by the well-trained reconstructor:

\vspace{-2mm}
\begin{equation}
 \mathbf{x}_{new}=R_{real}({\mathbf z}_{sem}^{sup},{\mathbf z}_{illu}),
\end{equation}

\noindent where ${\mathbf{z}_{sem}^{sup}} = E(\mathbf{t}^{tst})$ represent the extracted semantic features of support templates.


\newcommand{\var}[1]{\text{\texttt{#1}}}
\newcommand{\func}[1]{\text{\textsl{#1}}}

\makeatletter
\newcounter{phase}[algorithm]
\newlength{\phaserulewidth}
\newcommand{\setphaserulewidth}{\setlength{\phaserulewidth}}
\newcommand{\phase}[1]{%
  \vspace{-1.75ex}
  \Statex\leavevmode\llap{\rule{\dimexpr\labelwidth+\labelsep}{\phaserulewidth}}\rule{\linewidth}{\phaserulewidth}
  \Statex\strut\refstepcounter{phase}\textit{Phase~\thephase~--~#1}
  \vspace{-1.75ex}\Statex\leavevmode\llap{\rule{\dimexpr\labelwidth+\labelsep}{\phaserulewidth}}\rule{\linewidth}{\phaserulewidth}}
\makeatother

\setphaserulewidth{.7pt}

\begin{algorithm}[t]
  \caption{Sill-Net: illumination separation and augmentation}
  \label{algo:1}
  \begin{algorithmic}[]
    \State Training data: $\mathcal{X}=\left\{(\mathbf{x}_i, y_i, \mathbf{t}_i)\right\}_{i=1}^{N}$
    \vspace{0.02in}
    \State Test data: $\mathcal{X}^{tst}=\left\{(\mathbf{x}_i^{tst}, y_i^{tst}, \mathbf{t}_i^{tst})\right\}_{i=1}^{N^{tst}}$
    \vspace{0.02in}
    \State Models with learnable parameters: feature extractor $E$, classifier $P(y|\mathbf{z})$.
    \phase{Separation}
      \For{each training image pair $(\mathbf{x}_i,\mathbf{x}_j)$}
        \State Extract features: $\mathbf{z}_i = E(\mathbf{x}_i),\mathbf{z}_j = E(\mathbf{x}_j)$
        \State Split features: 
        \State $\mathbf{z}_i \rightarrow [\mathbf{{z}_{sem}}_{(i)}, \mathbf{{z}_{illu}}_{(i)}], \mathbf{z}_j \rightarrow [\mathbf{{z}_{sem}}_{(j)}, \mathbf{{z}_{illu}}_{(j)}]$
        \vspace{0.02in}
        \State Exchange mechanism for separation:
        \vspace{0.04in}
        \State $\mathcal{Z}_{exc}=\left\{ r{{\mathbf z}_{sem}}_{(i)} + (1-r) {{\mathbf z}_{illu}}_{(j)}, y_i \right\}$
        \vspace{0.04in}        
        \State $\mathcal{L}_{exc}=-\frac{1}{N_{exc}} \sum_{i=1}^{N_{exc}} \sum_{c=1}^{M} y_{ic} \log P\left(y_{ic}|{\mathbf z}_i \in \mathcal{Z}_{exc} \right)$
        \vspace{0.04in}
        \State Matching and reconstruction to the templates:
        \vspace{-0.05in}
        \begin{align*}
         & \mathcal{L}_{match}=\frac{1}{N} \sum_{i=1}^{N} \left( \mathcal{T} \left( {{\mathbf z}_{sem}}_{(i)}|_{{\mathbf x}_i} \right)-{{\mathbf z}_{sem}}_{(i)}|_{{\mathbf t}_i} \right )^2 \\
         & {
          \begin{aligned}
              \mathcal{L}_{recon} & = \frac{1}{N} \sum_{i=1}^{N} \sum_{j} -t_{ij} \log \hat{t}_{ij} \\ 
              & - \left(1-t_{ij}\right) \log \left(1-\hat{t}_{ij}\right)
          \end{aligned}}
        \end{align*}
        \vspace{-0.12in}
        \State Illumination constraint:
        \begin{equation}
          \begin{aligned}
            \mathcal{L}_{illu} & = -PIDA \\ &= -\sum_{c=1}^{M} \sum_{i=1}^{{N}_c}\mathcal{D} \left( \mathbb{E} \left({\mathbf z}_{illu}|_{\mathbf{x}_c, y_c} \right),{\mathbf z}_{illu}|_{{\mathbf x}_{ci}, y_c} \right) \nonumber
          \end{aligned}
        \end{equation}
        \State Total loss: $\mathcal{L}=\mathcal{L}_{exc} + \mathcal{L}_{match} + \mathcal{L}_{recon} + \mathcal{L}_{illu}$
      \EndFor
      \vspace{0.01in}
      \State Construct illumination repository: $\mathcal{Z}_{illu}=\left\{ {\mathbf{z}_{illu}}_{(i)} \right\}_{i=1}^{N}$
      \vspace{0.03in}
    \phase{Augmentation}
      \For{each support image $\mathbf{t}_i^{tst}$}
        \State Extract semantic features: ${\mathbf{z}_{sem}^{sup}}_{(i)} = E(\mathbf{t}_i^{tst})$
        \For {each illumination feature ${{\mathbf z}_{illu}}_{(j)}$}
          \State feature augmentation by mixup:
          \vspace{0.05in}
          \State $\mathcal{Z}_{aug} = \left\{ r {\mathbf{z}_{sem}^{sup}}_{(i)} + (1-r) {{\mathbf z}_{illu}}_{(j)}, y_i^{tst}\right\}$
          \vspace{0.05in}
          \State Train on the augmented feature set:
          \vspace{0.05in}
          \State $\mathcal{L}_{aug}=-\frac{1}{N_{aug}} \sum_{i=1}^{N} \sum_{c=1}^{M} y_{ic} \log P\left(y_{ic}|{\mathbf z}_i \in \mathcal{Z}_{aug} \right)$
        \EndFor
      \EndFor
    \phase{Inference}
      \For{each test image $\mathbf{x}_i^{tst}$}
        \vspace{0.05in}
        \State Extract features: ${\mathbf{z}}_{i} = E(\mathbf{x}_i^{tst})$
        \vspace{0.05in}
        \State Predict labels: $\hat{c}_i = \underset{c} {\arg \max } \  P \left(y_{ic}|{\mathbf z}_i \right)$
      \EndFor
  \end{algorithmic}
\end{algorithm}

\vspace{-2mm}
\section{Experiments}

\subsection{Datasets and experimental settings}

{\bf Datasets.} We first validate the effectiveness of Sill-Net on symbolic object datasets, i.e., four traffic sign datasets with various (especially extreme) illumination: German Traffic Sign Recognition Benchmark (GTSRB) \cite{stallkamp2012man}, Tsinghua-Tencent 100K (TT100K) \cite{zhu2016traffic}, Belgian Traffic Sign Classification (BTSC) \cite{mathias2013traffic} and Chinese Traffic Sign Database (CTSD) \footnote{Publicly available on \textit{http://www.nlpr.ia.ac.cn/pal/trafficdata/index.html}.}; three logo datasets, BelgaLogos \cite{joly2009logo, letessier2012scalable}, FlirckrLogos-32 \cite{romberg2011scalable} and TopLogo-10 \cite{su2017deep}, since these datasets contain various illumination. Table \ref{tab:dataset} shows the size and number of classes of each dataset.
We also evaluate the effectiveness of illumination feature augmentation on standardized one/few-shot image classification datasets: \emph{mini}ImageNet \cite{vinyals2016matching}, CUB \cite{wah2011caltech} and CIFAR-FS \cite{bertinetto2018meta}. Due to the page limit, we introduce the details of all datasets in Appendix B.

{\bf Evaluation tasks.} Generally, we evaluate our model by the following steps. 1) Utilize the training dataset (or subset) to separate out the illumination features. 2) The support samples are augmented with the illumination features to form an augmented feature set. 3) Train a classifier on the augmented feature set. 4) Prediction on the test dataset.

For symbolic object datasets, we validate our method on the following increasingly difficult classification tasks.

1) Traditional classification


We implement traditional classification tasks on symbolic object datasets under the regular splits of training and test (or validation) sets. We separate the semantic and illumination features for all training images. The illumination features are randomly augmented to the other semantic features, because the training images are also used as support samples here. We train the classifier on the augmented feature sets and then infer on the test sets.

2) One-shot classification

In this type of task, the training procedure requires no real-world images from test classes but one template image for each category. This task is similar to the one-shot classification.

For intra-dataset classification, we split the traffic sign datasets into training and test subsets with no class overlap. GTSRB is divided into 22 classes for training and another 21 classes for testing. For convenience, we denote this scenario by GTSRB$\rightarrow$GTSRB, where the training set is on the left side of the arrow and the test set is on the right. Similarly, we split BTSC and CTSD for one-shot classifications, denoted by BTSC$\rightarrow$BTSC and CTSD$\rightarrow$CTSD.

For cross-dataset evaluation, we train on large datasets and test on small datasets. We use GTSRB for training and TT100K for testing, denoted by GTSRB$\rightarrow$TT100K. And for logo classification, we use the largest BelgaLogos for training and the remaining two for testing respectively, denoted by Belga$\rightarrow$Flickr32 and Belga$\rightarrow$Toplogos.



3) Cross-domain one-shot classification

To further validate the generalization ability of our method, we perform a cross-domain one-shot evaluation by another two experiments, where the model is trained on traffic sign datasets and tested on logo datasets. Specifically, we train the model on GTSRB and test on FlickrLogos-32 and Toplogo-10. We denote these two scenarios as GTSRB$\rightarrow$Flickr32 and GTSRB$\rightarrow$Toplogos. The setup is more challenging compared to the previous scenarios, since we train the model in the domain of traffic sign datasets while testing in an entirely different domain of logo datasets.


For natural image datasets without regular templates such as ImageNet, the model can not be directly trained since our method requires template images for the separation of features. However, the repository with diverse illumination features separated from GTSRB can facilitate the classification of natural images via the proposed illumination feature augmentation. To verify the versatility of the illumination repository, we perform illumination feature augmentation on standardized few-shot classification datasets: \emph{mini}ImageNet \cite{vinyals2016matching}, CUB \cite{wah2011caltech}, and CIFAR-FS \cite{bertinetto2018meta}. The datasets are split into base classes, validation classes, and novel classes, following the settings in \cite{hu2020leveraging}. During training on the base classes, we augment the input images with the illumination features through mixup. The mixup proportion $r$ is set to $0.9$. Technical details are described in Appendix A.2 and C.3 due to the page limit. Here are two settings of few-shot classification: 

1) Transductive setting

We implement illumination feature augmentation during training the backbone of Wide residual networks (WRN) \cite{zagoruyko2016wide} with the pre-trained model provided by \cite{hu2020leveraging}. Once the backbone is trained, we extract the features of novel classes and then implement Power Transform and Maximum A Posteriori (PT+MAP) proposed by \cite{hu2020leveraging}. The results are presented in Section \ref{sec:transductive}

2) Inductive setting

In the inductive setting, we evaluate the effectiveness of feature illumination augmentation on common baselines in few-shot classification including Baseline, Baseline++ \cite{chen2019closer}, MAML \cite{finn2017model}, ProtoNet \cite{snell2017prototypical} and MatchingNet \cite{vinyals2016matching}. The results are presented in Section \ref{sec:inductive}.

{\bf Template image processing.} Previous studies \cite{tabelini2020deep} have shown that basic image processing on template images (as support samples) helps the network's generalization. In our experiment, we diversify the template images themselves using the following methods: geometric transformations (including translation, rotation and scaling), image enhancement (including brightness, color, contrast and sharpness adjustment), and blur. The template images are diversified and thus allow the model to learn more generalizable features. We observe that basic processing on template images improves model performance.

\vspace{-2mm}
\subsection{Experiments on symbolic object datasets} \label{sec:symbolic}

\subsubsection{Traditional classification} \label{sec:traditional}

\begin{table}[htbp]
 \caption{Symbolic object datasets with their traditional classification results ($\%$)}
 \vspace{-4mm}
 \label{tab:dataset}
 \scriptsize
 \begin{center}
   \setlength{\tabcolsep}{1.8mm}{
   \begin{tabular}{l|cccc|ccc}
    \toprule
      &\multicolumn{4}{|c|}{Traffic sign} & \multicolumn{3}{r}{Logo \qquad\qquad\qquad\qquad}\\
           {Dataset} & GTSRB  & TT100K & BTSC  & CTSD   & Belga   & Flickr32 & TopLogo \\
    \midrule
    Size     & 51839  & 11988  & 7095  & 6164   & 9585    & 3404     & 848     \\
    Classes  & 43     & 36     & 62    & 58     & 37      & 32       & 11      \\
    \midrule
    \midrule
    CNN3ST   & 99.71  & 99.06  & 98.87 & 83.55  & 87.68   & 94.78    & 85.06   \\
    Sill-Net & 99.68  & 99.53  & 98.97 & 97.19  & 89.48   & 95.80    & 89.66   \\
    \bottomrule
   \end{tabular}}
 \end{center}
\end{table}

The classification results on symbolic object datasets are shown in Table \ref{tab:dataset}. We compare our approach Sill-Net with the SOTA method, i.e., CNN with spatial transformers (CNN3ST) \cite{arcos2018deep}. Our method outperforms CNN3ST on most traffic sign and logo datasets, especially the Chinese Traffic Sign Dataset (CTSD). The results also indicate that Sill-Net performs better when training samples are insufficient.

\subsubsection{One-shot classification}

We compare our method with Siamese networks \cite{koch2015siamese} (SiamNet), Quadruplet networks \cite{kim2017co} (QuadNet), Matching
networks \cite{vinyals2016matching} (MatchNet) and Variational Prototyping-Encoder \cite{kim2019variational} (VPE) for one-shot classification, reported in Table \ref{tab:traffic} and \ref{tab:logo}. We quote or reproduce the results of the compared methods under their optimal settings, that is, VPE is implemented with augmentation and spatial transformer (VPE+aug+stn version), SiamNet and MatchNet is implemented with augmentation, while QuadNet is without augmentation. As shown in the tables, our method outperforms comparative methods in all scenarios.

\begin{table}[htbp]
 \caption{One-shot classification accuracies ($\%$) on traffic sign datasets. The best results are marked in bold.}
 \vspace{-2mm}
 \label{tab:traffic}
 \scriptsize
 \begin{center}
  \setlength{\tabcolsep}{5.5mm}{
   \begin{tabular}{lcc}
    \toprule
                                                  & GTSRB$\rightarrow$GTSRB          & GTSRB$\rightarrow$TT100K         \\
    No. support set                               & (22+21)-way                      & 36-way                           \\
    \midrule
    SiamNet \cite{koch2015siamese}                & 33.62                            & 22.74                            \\
    QuadNet \cite{kim2017co}                      & 45.2                             & N/A                              \\
    MatchNet \cite{vinyals2016matching}           & 53.30                            & 58.75                            \\
    VPE \cite{kim2019variational}                 & 83.79                            & 71.80                            \\
    Sill-Net                                      & \textbf{97.60}                   & \textbf{95.59}                   \\
    Sill-Net w/ NN                                & 94.53                            & 56.09                            \\
    \midrule
    \midrule
                                                  & BTSC$\rightarrow$BTSC            & CTSD$\rightarrow$CTSD            \\
    No. support set                               & (21+41)-way                           & (23+35)-way                           \\
    \midrule
    QuadNet \cite{kim2017co}                      & 51.92                            & 46.47                            \\
    VPE \cite{kim2019variational}                 & 78.46                            & 72.08                            \\
    Sill-Net                                      & \textbf{93.25}                   & \textbf{93.53}                   \\
    Sill-Net w/ NN                                & 89.67                            & 69.43                            \\
    \bottomrule
   \end{tabular}}
 \end{center}
\vspace{-4mm}
\end{table}

In traffic sign classification, Sill-Net outperforms the second best method VPE by a large margin of 13.81$\%$ (from 83.79$\%$ to 97.60$\%$) in GTSRB$\rightarrow$GTSRB, 23.79$\%$ (from 71.80$\%$ to 95.59$\%$) in GTSRB$\rightarrow$TT100K, 21.45$\%$ (from 72.08$\%$ to 94.53$\%$) in CTSD$\rightarrow$CTSD, and 14.79$\%$ (from 78.46$\%$ to 93.25$\%$) in BTSC$\rightarrow$BTSC (see Table \ref{tab:traffic}). It indicates that training on the features augmented by illumination information does help the real-world classification, even though only one template image is provided. It is notable that in the cross-dataset scenario GTSRB$\rightarrow$TT100K, Sill-Net achieves a comparable performance to the intra-dataset scenario GTSRB$\rightarrow$GTSRB, while VPE performs much worse in the cross-dataset scenario. We surmise it is because VPE learns latent embeddings generalizable to test classes in the same domain (GTSRB), but the generalization might be discounted when the target domain is slightly shifted (from GTSRB to TT100K). It is observed that the illumination conditions in GTSRB are quite similar to that in TT100K, therefore Sill-Net shows better generalization performance by making full use of the illumination information in GTSRB.

\begin{table}[htbp]
 \caption{One-shot classification accuracy ($\%$) on logo datasets. The best results are marked in bold.}
 \vspace{-1mm}
 \label{tab:logo}
 \scriptsize
 \begin{center}
  \setlength{\tabcolsep}{6mm}{
   \begin{tabular}{lcc}
    \toprule
                                                  & Belga$\rightarrow$Flickr32       & Belga$\rightarrow$Toplogos       \\
    No. support set                               & 32-way                           & 11-way                           \\
    \midrule
    SiamNet \cite{koch2015siamese}                & 22.82                            & 30.46                            \\
    QuadNet \cite{kim2017co}                      & 37.72                            & 36.62                            \\
    MatchNet \cite{vinyals2016matching}           & 40.95                            & 35.24                            \\
    VPE \cite{kim2019variational}                 & 53.53                            & 57.75                            \\
    Sill-Net                                      & {\textbf{65.21}}                 & {\textbf{84.43}}                 \\
    Sill-Net w/ NN                                & 45.62                            & 39.75                            \\
    \bottomrule
   \end{tabular}}
 \end{center}
\vspace{-1mm}
\end{table}

In logo classification, Sill-Net improves the performance by 11.68$\%$ (from 53.53$\%$ to 65.21$\%$) and 26.68$\%$ (from 57.75$\%$ to 84.43$\%$) compared to VPE, respectively in two scenarios (see Table \ref{tab:logo}). The improvement of accuracies in logo classification is not comparable to that in traffic classification, which is due to the undesirable quality of the training logo dataset. The GTSRB is the largest dataset with various illumination conditions. And the traffic signs are always complete and well localized in the images so that illumination features can be separated more easily. In contrast, the separation is harder for logo dataset due to incomplete logos, color changes, and non-rigid deformation (e.g., logos on the bottles).

As introduced before, a parametric classifier is trained on the augmented feature set in our method. For comparison, we also implement Sill-Net with the nearest-neighbor (NN) classifier, a non-parametric approach commonly used in few-shot classification. The semantic features of test images are extracted by the same separation network and then classified by NN. The results (see Sill-Net w/ NN in Tables \ref{tab:traffic} and \ref{tab:logo}) show that training a parametric classifier with feature augmentation performs better than classifying the semantic features by NN, especially when the separation of semantic features is not good enough due to domain shift (GTSRB$\rightarrow$TT100K) or insufficient training samples (CTSD$\rightarrow$CTSD and BTSC$\rightarrow$BTSC). 

\subsubsection{Cross-domain one-shot classification}

Sill-Net achieves the best results among all methods in cross-domain one-shot classification tasks, as shown in Table \ref{tab:cross}. It outperforms VPE by a large margin of 23.63$\%$ (69.75$\%$ compared to 46.12$\%$) in GTSRB$\rightarrow$Flickr32 and 39.86$\%$ (69.46$\%$ compared to 29.60$\%$) in GTSRB$\rightarrow$Toplogos.

\begin{table}[htbp]
 \caption{Cross-domain one-shot classification accuracy ($\%$). The models are trained on the traffic sign dataset (GTSRB) and tested on the logo datasets. The best results are marked in bold.}
 \vspace{-1mm}
 \label{tab:cross}
 \scriptsize
 \begin{center}
  \setlength{\tabcolsep}{5.5mm}{
   \begin{tabular}{lcc}
    \toprule
                                                  & GTSRB$\rightarrow$Flickr32       & GTSRB$\rightarrow$Toplogos       \\
    No. support set                               & 32-way                           & 11-way                           \\
    \midrule
    QuadNet \cite{kim2017co}                      & 28.41                            & 25.38                            \\
    VPE \cite{kim2019variational}                 & 46.12                            & 29.60                            \\
    Sill-Net                                      & {\textbf{69.75}}                 & {\textbf{69.46}}                 \\
    Sill-Net w/ NN                                & 38.94                            & 25.47                            \\
    \bottomrule
   \end{tabular}}
 \end{center}
\vspace{-1mm}
\end{table}

The results illustrate that our method is still generalizable when the domain is transferred from traffic signs to logos. The unsatisfactory results of VPE are predictable. VPE learns a generalizable similarity embedding space of the semantic information among the same or similar domain (i.e., from traffic signs to traffic signs or from logos to logos). However, the embeddings learned from traffic signs are difficult to generalize to logos. In contrast, our method learns well-separated semantic and illumination representation and the illumination features are generalizable for augmentation to the template images from novel domains.

\subsubsection{Ablation study}

In this section, we delve into the contribution of each component of our method. The components under evaluation include the exchange mechanism, the matching and reconstruction module, the illumination constraint, and template image processing, as shown in Table \ref{tab:ablation}. We disable one component at a time and then record the performance to assess its importance. The experiments are implemented in the one-shot classification scenario GTSRB$\rightarrow$GTSRB.

The results demonstrate that the exchange mechanism and matching module are the core components of our method. The accuracy of the model drops to 48.10$\%$ without the exchange mechanism. It is because the semantic and illumination features cannot be well separated without the exchange mechanism. The remaining semantic information in the illumination features is useless, or even would interfere with the recognition when they are combined with the semantic features of other objects during feature augmentation, hurting the performance of the model.

Meanwhile, the matching module cooperating with the separation module further separates the semantic and illumination features. The matching module corrects the deformation of the object features. It retains the concrete semantic information (e.g., the outline of objects and semantic details of object contents) with the supervision of template images. Without the matching module, the semantic features would be less informative, so the separation module would have difficulty separating the illumination features from the semantic features. Therefore, the accuracy of the model drops to 54.27$\%$ when the matching module is removed.

\begin{table}[htbp]
 \begin{center}
 \caption{Ablation study results ($\%$) in the one-shot classification scenario GTSRB$\rightarrow$GTSRB. We disable one component at a time and record the performance of Sill-Net.}
 \label{tab:ablation}
 \small
  \begin{tabular}{p{4cm}l}
   \toprule
   \textbf{Factor}             & \textbf{Accuracy (decrement)} \\
   \midrule
   w/o exchange mechanism      & \qquad \qquad 48.10 (-49.50)  \\
   w/o matching module         & \qquad \qquad 54.27 (-43.33)  \\
   w/o reconstruction module   & \qquad \qquad 80.74 (-16.86)  \\
   w/o illumination constraint & \qquad \qquad 90.73 (-6.87)   \\
   w/o template processing     & \qquad \qquad 80.19 (-17.41)  \\
   \midrule
   full method                 & \qquad \qquad 97.60           \\
   \bottomrule
  \end{tabular}
 \end{center}
\vspace{-4mm}
\end{table}

The accuracy decreases by 16.86$\%$ without the reconstruction module, which also strives to make semantic features more informative. The matching module helps the model capture some level of the concrete semantic information, while the reconstruction module prompts to retain delicate details of the object.

The illumination constraint increases the accuracy by 6.87$\%$. Intuitively, it reduces the semantic information in illumination features and thus enhances their quality. Higher-quality illumination representation improves the effectiveness of our feature augmentation method, which is consistent with the results.

Furthermore, template image processing improves model performance as expected. The processing methods (i.e., geometric transformations, image enhancement, and blur as introduced before) diversify the template images so that the trained model is more generalizable. Under the combined effect of the proposed illumination augmentation in the feature space and the variation of template images, the full model achieves the best results among the existing methods.


\subsubsection{Comparison with standard augmentations}

In this section, we thoroughly compare our method with standard augmentation techniques. We maintain the extraction network in the separation phase, while during the augmentation phase, we apply standard augmentation to the semantic features instead of illumination feature augmentation. The results of GTSRB$\rightarrow$GTSRB are shown in Table \ref{tab:standard aug}.

\begin{table}[htbp]
 \caption{The results ($\%$) of comparison with standard augmentations in GTSRB$\rightarrow$GTSRB}
 \vspace{-2mm}
 \label{tab:standard aug}
 \begin{center}
   \begin{tabular}{lcccc}
    \toprule
    Method      & Warping      & Cropping     & Rotation     & Flipping     \\
    \midrule
    Accuracy    & 41.11        & 31.14        & 26.60        & 21.61        \\
    \midrule
    Method      & Enhancement  & Blur         & All-Aug      & Illu-Aug     \\
    \midrule
    Accuracy    & 34.10        & 34.84        & 53.44        & 80.19        \\
    \bottomrule
   \end{tabular}
 \end{center}
\end{table}

For geometric transformations, we apply random warping, cropping, rotation, and flipping. Image enhancement (random brightness, color, contrast and sharpness adjustments) and blur are also implemented for comparison. We also apply all above standard augmentation techniques together (All-Aug) to achieve better results. To be fair, the template processing (including standard augmentations) of our method (Illu-Aug) is removed.

As we can see, the results of one single standard augmentation are not satisfactory. It is because a single augmentation is always difficult to cover all the various deformations. The results are improved when applying all standard augmentations together. However, our illumination feature augmentation still outperforms the combination of these augmentations.

\subsubsection{Visualization}\label{sec:visualization}

\begin{figure*}[htbp]
 \vspace{-4mm}
 \begin{center}
  \includegraphics[width=1.0\textwidth]{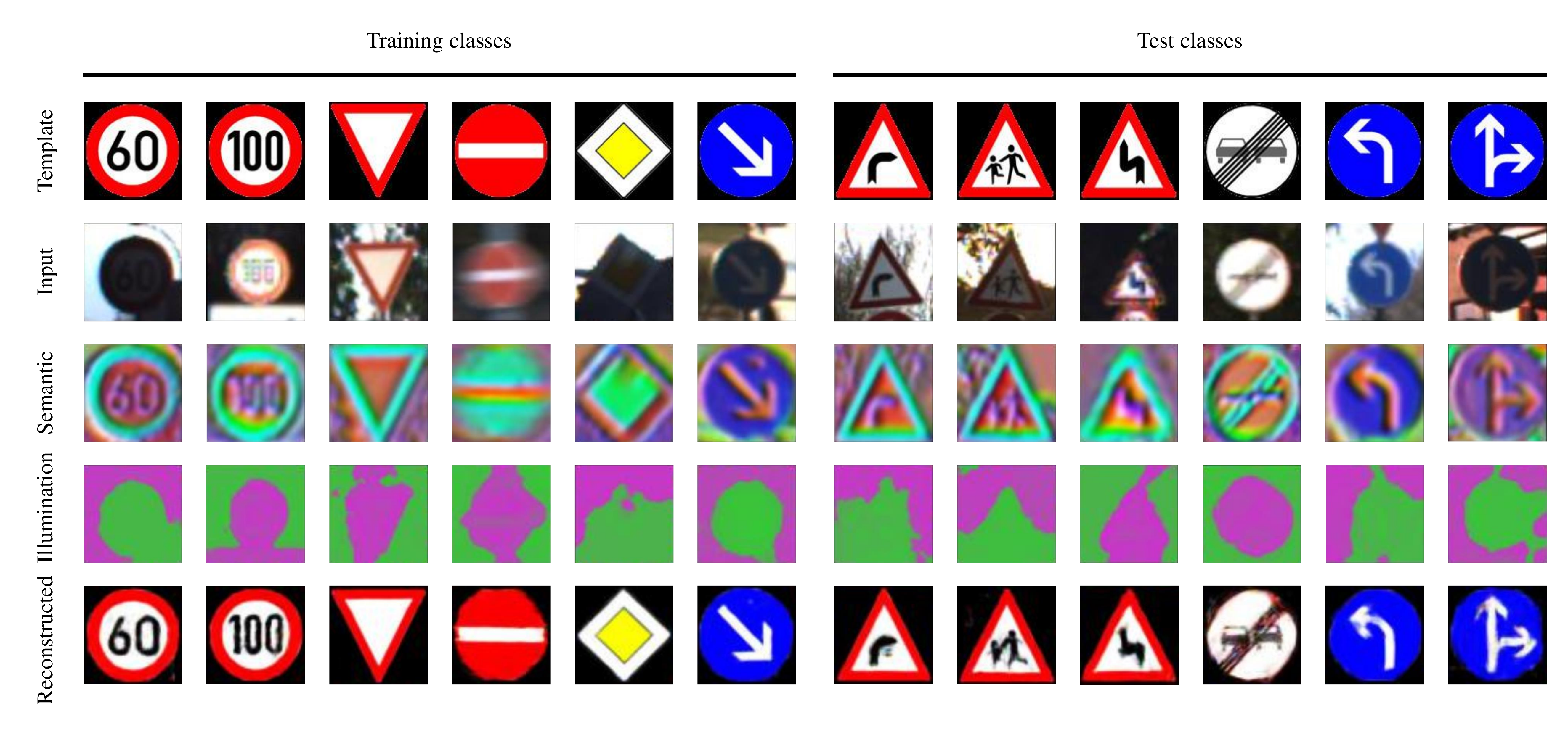}
 \end{center}
 \vspace{-8mm}
 \caption{Visualization of the separated features and the reconstructed template images from training and test classes. The first two rows show the input images and their corresponding template images. The third and fourth rows show the semantic and illumination features of the input images separated by our model. The last row shows the template images reconstructed from the semantic features. Note that the training and test datasets share no common classes.}
 \label{fig:visualization}
\end{figure*}

{\bf Feature visualization.} Figure \ref{fig:visualization} shows the separated semantic and illumination features of the images from training and test classes in GTSRB, visualized in the third and fourth lines. Note that the training and test datasets share no common classes. As shown in the figure, the semantic features delicately retain information consistent with the template images for both training and test classes. It is due to three aspects. First, the extractor maintains the size and spatial information of the features. Second, although objects in the input images vary in size and position, the features are corrected to the normal situation corresponding to the template images via the spatial transformer in the matching module. Third, the reconstruction module  promotes the semantic feature to retain the details of the objects.


In contrast, the semantic information is effectively reduced in illumination features. They reflect the illumination conditions in the original images to a certain extent. Intuitively, the pink parts in the features represent the bright illumination while the green parts represent the dark illumination. Such well-separated representation lays the foundation for the good performance of our model.


{\bf Template reconstruction.} While the reconstructor serves to obtain informative semantic features during training, it can also retrieve the template images in the inference phase. As shown in the last row of Figure \ref{fig:visualization}, the reconstructor robustly generates the template images of both the training and test samples, regardless of illumination variance, object deformation, blur, and low resolution of the images. Not only outlines of the symbol contents but also fine details are well restored in the generated template images, which improves the reconstruction results by VPE. Our results further demonstrate that the proposed model has learned good representation of semantic information for both classification and reconstruction.

%
%
%

\vspace{-2mm}
\subsection{Experiments on the illumination repository}

\subsubsection{Illumination repository expansion performance}\label{sec:expansion}

We have constructed a repository containing 51,839 illumination features separated from the training subset of GTSRB. As we observed that the repository includes similar illumination features that might be redundant and less useful for training, we select the representative samples to reduce the number while keeping the important structure of illumination features. Subsequently, we expand the illumination feature space through linear interpolation to further diversify the augmentation features (see Section \ref{sec:repository}). 

{\bf Feature selection} We first select the illumination features by $k$-means clustering. The features in the repository are clustered into $k$ partitions. We gather the cluster center of each partition to form a compact repository retaining representative illumination features. In the experiment, we tried both $k=100$ and $k=1000$. 

{\bf Feature interpolation}
We then expand the feature space using the selected features in the compact repository. We generate new illumination features through linear interpolation of random pairs of the selected features. The number of the generated features are set as $N_{exp}=40000$. We evaluate Sill-Net by augmenting the support samples with the generated illumination features in the scenario GTSRB$\rightarrow$GTSRB.

The results are shown in Table \ref{tab:expansion}. As we can see, feature selection can greatly reduce the size of the illumination repository (from 51,839 to $k=100$ or $k=1000$) while only causing a small decrease in accuracy, indicating that the raw illumination features are redundant to a certain extent. 

Meanwhile, certain expansion of the feature space can generate useful novel illumination features, which can be used as feature augmentation to improve model performance. In our experiment, the linear interpolation method can enhance the accuracy from 85.36\% to 91.12\% in the case that $k=100$. When $k=1000$, the accuracy increases from 90.40\% to 97.62\%, even outperforming the best result (97.60\%) achieved by the raw illumination features.

\begin{table}[htbp]
 \vspace{-2mm}
 \caption{One-shot classification accuracy ($\%$) when augmenting the support samples with generated illumination features from $k$ selected samples in the scenario GTSRB$\rightarrow$GTSRB. The best result is marked in bold.}
 \vspace{-2mm}
 \label{tab:expansion}
 \small
 \begin{center}
  \setlength{\tabcolsep}{8mm}{
   \begin{tabular}{lcc}
    \toprule
    $k$                       & Selection                    & Interpolation                              \\
    \midrule
    100                       & 85.36                        & 91.12                                      \\
    1000                      & 90.40                        & {\textbf{97.62}}                           \\
    51839                     & 97.60                        & N/A                                        \\
    \bottomrule
   \end{tabular}}
 \end{center}
\vspace{-4mm}
\end{table}


\begin{figure}[htbp]
\vspace{-2mm}
 \begin{center}
 \vspace{-4mm}
  \hspace{-9mm}\includegraphics[width=1.1\columnwidth]{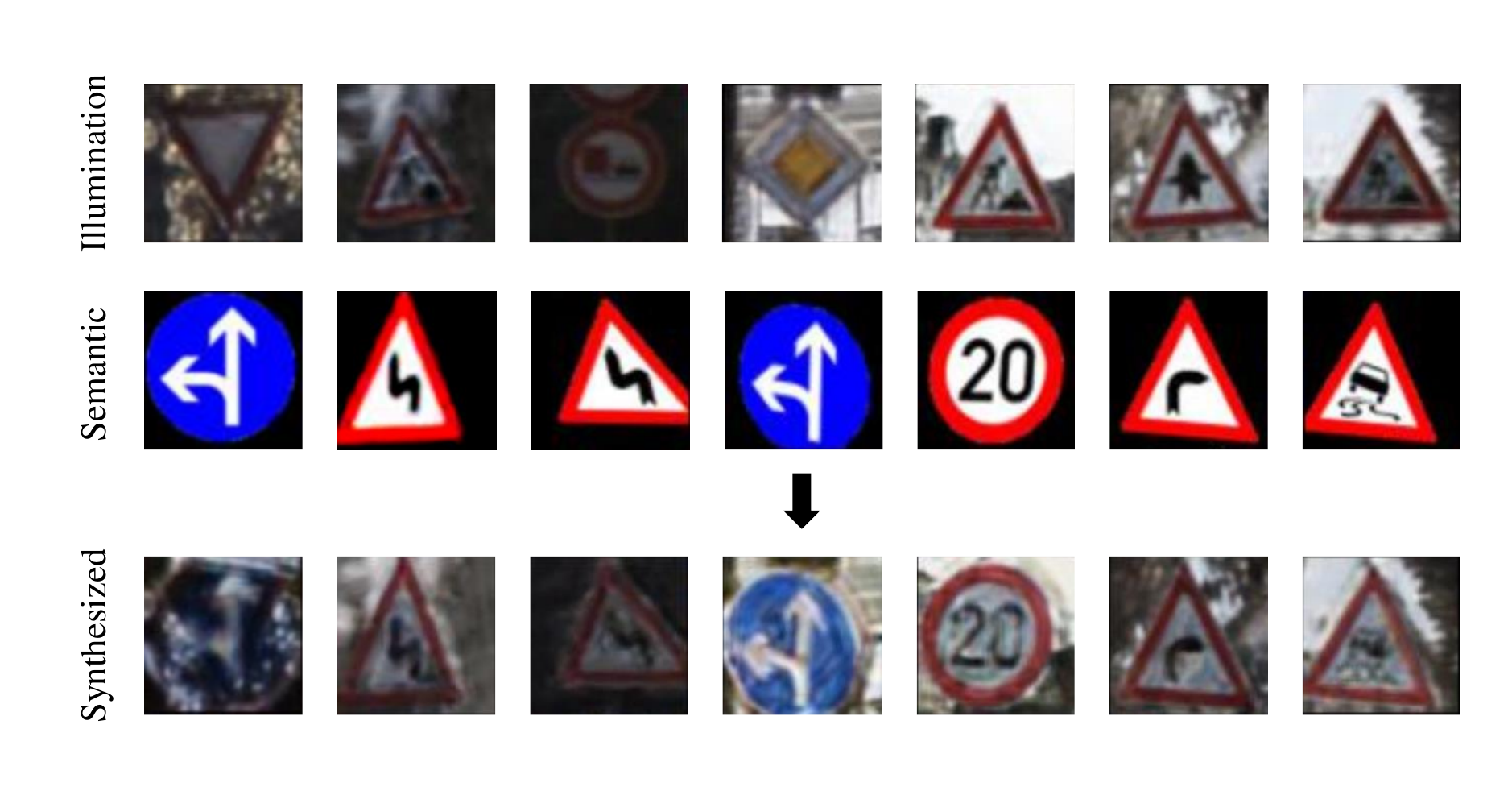}
 \end{center}
 \vspace{-6mm}
 \caption{Illumination transplantation. The synthesized images show objects of support samples under illumination separated from training images.}
 \label{fig:transplantation}
\end{figure}

\subsubsection{Illumination reconstruction and transplantation}\label{sec:transplantation}

To further interpret the separated illumination features intuitively, we reconstructed them back to visual images as described in Section \ref{sec:transplantation}. Figure \ref{fig:transplantation} shows the results of transplanting the illumination features separated from training images to the support template images. As we can see, the reconstructed illumination condition on new objects is similar to that of original images, while the semantic information is completely replaced in the synthesized images. Fine details of illumination such as delicate light and shadows can be well retrieved from the features. The results indicate that the illumination features indeed contain retrievable information related to illumination conditions on both foreground and background.

\subsection{Generalized one/few-shot classification on natural image datasets} \label{sec:few-shot}

\begin{table}[htbp]
 \vspace{-4mm}
 \caption{Generalized 1-shot and 5-shot classification accuracy ($\%$) on \emph{mini}ImageNet, CUB and CIFAR-FS in the transductive setting. Our method is aligned with the pipeline of PT+MAP with the backbone of WRN, trained with the images augmented by the illumination repository. The best results are marked in bold.}
 \label{tab:transductive}
 \scriptsize
 \begin{center}
 \setlength{\tabcolsep}{3.5mm}{
   \begin{tabular}{llcc}
    \toprule
                                             &                    & \multicolumn{2}{c}{\emph{mini}ImageNet}           \\
    Method                                   & Backbone           & 1-shot                 & 5-shot                   \\
    \midrule
    BD-CSPN \cite{liu2019prototype}          & WRN                & 70.31 $\pm$ 0.93       & 81.89 $\pm$ 0.60         \\
    Transfer+SGC \cite{hu2020exploiting}     & WRN                & 76.47 $\pm$ 0.23       & 85.23 $\pm$ 0.13         \\
    TAFSSL \cite{lichtenstein2020tafssl}     & DenseNet121        & 77.06 $\pm$ 0.26       & 84.99 $\pm$ 0.14         \\
    DFMN-MCT \cite{kye2020transductive}      & ResNet12           & 78.55 $\pm$ 0.86       & 86.03 $\pm$ 0.42         \\
    PT+MAP \cite{hu2020leveraging}           & WRN                & 82.92 $\pm$ 0.26       & 88.82 $\pm$ 0.13         \\
    Illu-Aug (ours)                           & WRN                & {\textbf{82.99} $\pm$ \textbf{0.23}}    & {\textbf{89.14} $\pm$ \textbf{0.12}} \\

    \toprule
                                              &                    & \multicolumn{2}{c}{CUB}                           \\
    Method                                    & Backbone           & 1-shot                 & 5-shot                   \\
    \midrule
    BD-CSPN \cite{liu2019prototype}           & WRN                & 87.45 \qquad\quad\quad & 91.74 \qquad\quad\quad   \\
    Transfer+SGC \cite{hu2020exploiting}      & WRN                & 88.35 $\pm$ 0.19       & 92.14 $\pm$ 0.10         \\
    PT+MAP \cite{hu2020leveraging}            & WRN                & 91.55 $\pm$ 0.19       & 93.99 $\pm$ 0.10         \\
    Illu-Aug (ours)                            & WRN                & {\textbf{94.73} $\pm$ \textbf{0.14}}    & {\textbf{96.28} $\pm$ \textbf{0.08}} \\

    \toprule
                                              &                    & \multicolumn{2}{c}{CIFAR-FS}                      \\
    Method                                    & Backbone           & 1-shot                 & 5-shot                   \\
    \midrule
    DSN-MR \cite{simon2020adaptive}           & ResNet12           & 78.00 $\pm$ 0.90       & 87.30 $\pm$ 0.60         \\
    Transfer+SGC \cite{hu2020exploiting}      & WRN                & 83.90 $\pm$ 0.22       & 88.76 $\pm$ 0.15         \\
    PT+MAP \cite{hu2020leveraging}            & WRN                & 87.69 $\pm$ 0.23       & 90.68 $\pm$ 0.15         \\
    Illu-Aug (ours)                           & WRN                & {\textbf{87.73} $\pm$ \textbf{0.22}}    & {\textbf{91.09} $\pm$ \textbf{0.15}} \\

    \bottomrule

   \end{tabular}}
 \end{center}
 \vspace{-4mm}
\end{table}

To verify the versatility of the illumination features extracted from GTSRB, we conduct feature augmentation with our illumination repository on standardized few-shot classification benchmarks (i.e., \emph{mini}ImageNet, CUB, and CIFAR-FS) and compare the performance with other state-of-the-art approaches. 

\subsubsection{Transductive setting} \label{sec:transductive}

We augment the training images with the illumination features extracted from GTSRB when training the backbone and implement Power Transformation and Maximum A Posteriori (PT+MAP) proposed by \cite{hu2020leveraging}. The results are shown in Table \ref{tab:transductive} (our method is denoted by Illu-Aug).


As we can see, our method achieves state-of-the-art performance on all benchmarks for both 1-shot and 5-shot classification. Generally, our method performs better when the objects are not deformed much compared to the representative template images, so that it achieves the most improvements on the CUB dataset. Also, providing more representative template images (i.e., increase the diversity of template images) to be augmented improves the performance. So our method achieves more improvements in 5-shot classification compared to 1-shot classification.

\subsubsection{Inductive setting} \label{sec:inductive}

\begin{table}[htbp]
 \vspace{-2mm}
 \caption{Generalized 1-shot and 5-shot classification accuracy ($\%$) on CUB in the inductive setting. We augment training images with the illumination repository during the training of baselines. The best results are marked in bold.}
 \vspace{-2mm}
 \label{tab:inductive}
 \scriptsize
 \begin{center}
 \setlength{\tabcolsep}{1.0mm}{
   \begin{tabular}{llcccc}
    \toprule
                 &           & \multicolumn{2}{c}{1-shot}                                                  &\multicolumn{2}{c}{5-shot}          \\
    Method       & Backbone  & Standard                             & Illu-Aug                             & Standard                               & Illu-Aug                                    \\
    \midrule
    Baseline     & Conv4     & {\textbf{47.12} $\pm$ \textbf{0.74}} & 46.84 $\pm$ 0.79                     & 64.16 $\pm$ 0.71                       & {\textbf{66.58} $\pm$ \textbf{0.69}}        \\
    Baseline++   & Conv4     & 60.53 $\pm$ 0.83                     & {\textbf{62.83} $\pm$ \textbf{0.87}} & 79.34 $\pm$ 0.61      
& {\textbf{79.74} $\pm$ \textbf{0.60}}                                 \\
    MAML         & ResNet10  & 70.32 $\pm$ 0.99                     & {\textbf{72.45} $\pm$ \textbf{0.91}} & 80.93 $\pm$ 0.71                       & {\textbf{82.58} $\pm$ \textbf{0.64}}        \\
    MatchingNet  & ResNet18  & {\textbf{73.49} $\pm$ \textbf{0.89}} & 73.26 $\pm$ 0.90                     & 84.45 $\pm$ 0.58                       & {\textbf{85.18} $\pm$ \textbf{0.60}}        \\
    ProtoNet     & ResNet18  & 72.99 $\pm$ 0.88                     & {\textbf{74.35} $\pm$ \textbf{0.91}} & 86.64 $\pm$ 0.51                       & {\textbf{87.80} $\pm$ \textbf{0.48}}        \\
    
    \bottomrule  
   \end{tabular}}
 \end{center}
\vspace{-4mm}
\end{table}

In the inductive setting, we apply feature illumination augmentation to common few-shot baselines and compare them with standard implementations. We evaluate on CUB since our method improves the most on this dataset in the transductive setting. The results show that generally our method performs better in 5-shot classification. It does not always perform well on 1-shot tasks because illumination feature augmentation can not address the variation of objects themselves, so more support samples (e.g., 5-shot) are needed. For Baseline++, MAML and ProtoNet, our method helps improve the performance in both 1-shot and 5-shot classifications.

Overall, the results demonstrate that the repository with diverse illumination features separated from GTSRB can facilitate the generalized one/few-shot classification on natural images in both the transductive and inductive settings via our proposed augmentation method.


\section{Discussions}

Unlike most previous works concentrating on how to augment the semantic part of features \cite{bai2020decaug,li2022semantic,wang2021regularizing} or learn the augmentation policies \cite{cubuk2019autoaugment,cubuk2020randaugment}, we provide a novel perspective to remove the semantic part of images and store the remaining to the repository for augmentations in feature space. Our method can be widely applied to a series of training scenarios. In the case that the training samples with certain illumination conditions are limited in the dataset, we can augment these samples with that type of illumination features separated from other images (or simply use the illumination features in our repository). Furthermore, we can utilize the method to expand a few support samples or even only one (e.g., template images) to form a large training dataset, solving the problem of lacking annotated real-world data. Overall, the imbalance both in size and illumination conditions of the dataset could be alleviated since we can transplant illumination information to specific training samples limited in number or insufficient in illumination diversity.

However, some issues and limitations need to be clarified.

First, why do we blend illumination features during inference? Since we mix semantic and illumination features during augmentation to train the classifier, we have to maintain the same input pattern for classification during inference. The accuracy will decrease if we classify semantic features without blending during inference because the distribution of input features to the classifier is shifted from that of mixed features training the classifier \cite{carratino2020mixup}. So we must maintain the same input pattern for the classifier during training and inference to obtain optimal results. Further investigation of input patterns is shown in Appendix C.2. 


Second, there is no sufficient supervision or constraints to ensure that the separated illumination features contain only illumination information. One image contains various information, including object semantics referring to labels, light and shadows, irrelevant background, etc. And the illumination information is sometimes related to background objects, such as the sky, lamps and trees. It is difficult to disentangle all these factors. Theoretically, existing constraints only guarantee that the separated illumination features are non-semantic. So we have to carefully choose the training datasets with various illumination and few irrelevant objects. Experiments in Section \ref{sec:visualization} (Figure \ref{fig:visualization}) show that the visualized illumination features mainly reflect the illumination intensity. And the illumination reconstruction results in Section \ref{sec:transplantation} (Figure \ref{fig:transplantation}) show that these features are influenced by general illumination such as the light from the sky or shadows on trees. Since we conduct experiments to verify the effectiveness of features mostly reflecting illumination, we discreetly call them "illumination features" rather than the broader concept "non-semantic features".


Third, the training of illumination separation is limited to simple object datasets with various illumination. It is not suitable for natural images with complex structures because during separation, object features have to be spatially transformed before being matched and reconstructed to regular templates. More sophisticated methods such as semantic transformations \cite{wang2021regularizing} should be investigated to solve non-spatial transformations for future research. 

Finally, another limitation is the assumption that the illumination distribution learned from training data should characterize that of test data, which is not always fully satisfied. To better fulfill this assumption, we have made the illumination repository representative and diverse through feature selection and interpolation, achieving satisfactory results for both symbolic objects (Section \ref{sec:expansion}) and natural images (Section \ref{sec:few-shot}). Future research on the repository is needed to make our method more generalizable.

\vspace{-2mm}
\section{Related works}

\textbf{Data augmentation.} Deep learning is always heavily reliant on large amounts of data to avoid the overfitting of the networks. Data augmentation is an effective data-space solution to the problem of limited data \cite{shorten2019survey}. It includes a series of methods that artificially inflate the size of the training dataset through either data warping or oversampling. These augmentations can be divided into basic image manipulations such as geometric transformations \cite{taylor2018improving}, color transformations \cite{mikolajczyk2018data}, random erasing \cite{zhong2020random}, noise injection \cite{moreno2018forward}, image mixing \cite{inoue2018data}, kernel filters \cite{kang2017patchshuffle}, etc., and deep learning approaches including feature space augmentations \cite{devries2017dataset}, adversarial training \cite{antoniou2017data}, neural style transfer \cite{zheng2019stada} and GAN-based data augmentations \cite{bowles2018gan}. Recently, many learnable augmentation methods \cite{cubuk2019autoaugment,cubuk2020randaugment,perez2017effectiveness,luo2020learn,zoph2020learning,hauberg2016dreaming} have been proposed and applied to image classification \cite{perez2017effectiveness}, text recognition \cite{luo2020learn}, and objection detection \cite{zoph2020learning}. AutoAugment \cite{cubuk2019autoaugment} designs a search space to find the best policy consisting of different random augmentation sub-policies to yield the highest validation accuracy on a target dataset, and RandAugment \cite{cubuk2020randaugment} further reduces the complexity of augmentation strategy search and improve efficiency. Our Sill-Net uses a similar augmentation technique to image mixing \cite{inoue2018data}. It is also a learnable augmentation method but it directly learns the augmented feature space itself instead of the parameters of augmentation policies.

Feature space augmentation implements the transformation in a learned feature space rather than the input space. DeVries and Taylor \cite{devries2017dataset} first performs the transformation in the feature space by adding noise, interpolating, or extrapolating. Recently, augmentation methods in the semantic feature space are proposed to regularize deep networks \cite{wang2021regularizing,bai2020decaug,dvornik2019importance}. We usually copy and paste the learned semantic features into new scarce data \cite{chu2020feature,ghiasi2021simple}, or the statistics such as mean and variance of semantic features to increase diversity \cite{li2021feature}. Prediction on mixed labels of mixed data is also proven to be an effective way of augmentation \cite{wu2020generalization,olsson2021classmix}.   Unlike these methods, we augment the samples in an easy-to-use way without constraints on data distribution and in a more interpretable and intuitive perspective that the augmented features represent different illumination. We are the first to focus on removing the semantic part and storing the remaining illumination to the repository for augmentations while other works concentrate on how to augment the semantic part of features.



\textbf{Feature disentanglement.} One important issue of computer vision is to learn the disentangled representation comprising a number of latent factors, with each factor controlling an interpretable aspect of the generated data \cite{bengio2013representation}. Some works focus on disentanglement learning with respect to specified labels \cite{jha2018disentangling,bouchacourt2017multi}. They are proposed to extract features composed of two parts: one summarizes the specified factors of variation associated with the labels (content), while the other summarizes the remaining unspecified variability (e.g., style). One basic idea is to pair the content factor of a data sample with the style factor of another sample with the same label \cite{mathieu2016disentangling}. The resulting representation should be able to recover the latter sample using a decoder. Such an idea is also applied to a weakly-supervised setting that only members of a given group sharing the same label are available, whereas different groups are not required to represent different classes \cite{bouchacourt2017multi,hosoya2019group,nemeth2020adversarial}. Our study disentangles representation into semantic parts associated with labels as specified content and illumination parts as unspecified variability.


\textbf{Few-shot learning.} Compared to the common machine learning paradigm that involves large-scale labeled training data, the development of Few-Shot Learning (FSL) is relatively tardy due to its intrinsic difficulty. Early efforts for FSL were based on generative models that sought to build the Bayesian probabilistic framework \cite{fei2006one,lake2015human}. Recently, more attention was paid on meta-learning, which can be generally summarized into five sub-categories: \emph{learn-to-measure} (\emph{e.g.}, MatchNets \cite{vinyals2016matching}, ProtoNets \cite{snell2017prototypical}, RelationNets \cite{yang2018learning}), \emph{learn-to-finetune} (\emph{e.g.}, Meta-Learner LSTM \cite{ravi2016optimization}, MAML \cite{finn2017model}), \emph{learn-to-remember} (\emph{e.g.}, MANN \cite{santoro2016meta}, SNAIL \cite{mishra2018simple}), \emph{learn-to-adjust} (\emph{e.g.}, MetaNets \cite{Munkhdalai2017Meta}, CSN \cite{munkhdalai2018rapid}) and \emph{learn-to-parameterize} (\emph{e.g.}, DynamicNets \cite{gidaris2018dynamic}, Acts2Params \cite{qiao2018few}). In this work, we use tasks similar to one-shot learning to evaluate our method and compare the results to these FSL methods. Our method is partly a data augmentation strategy to solve the data scarcity problems, thus compatible with almost all FSL methods.

\section{Conclusion}

In this paper, we develop a novel neural network architecture named \textbf{S}eparating-\textbf{Ill}umination \textbf{Net}work (Sill-Net) to separate illumination features from images and then augment them to support samples. Extensive experiments validate the feasibility and effectiveness of illumination-based feature augmentation in different classification tasks. Our method not only outperforms the state-of-the-art (SOTA) methods by a large margin on symbolic object datasets but also achieves improvements on natural image benchmarks. It is a novel perspective to remove the semantic part of images and use the remaining for feature augmentations. Currently, the separation of semantic and illumination features is limited to simple-structured symbolic objects. In our future work, we shall adopt the separation phase of our method to natural images with complex structures.



%

%

\vspace{-4mm}

\ifCLASSOPTIONcompsoc
  \section*{Acknowledgments}
\else
  \section*{Acknowledgment}
\fi

This work was supported by the National Key Research and Development Program of China (No. 2018AAA0100701), Tsinghua-Toyota Joint Research Institute Cross-discipline Program, and Beijing Academy of Artificial Intelligence (BAAI).

\ifCLASSOPTIONcaptionsoff
  \newpage
\fi



%
\vspace{-4mm}

\bibliography{manuscript}
\bibliographystyle{IEEEtran}

%


\vspace{-11mm}
\begin{IEEEbiography}[{\includegraphics[width=1in,height=1.25in,clip,keepaspectratio]{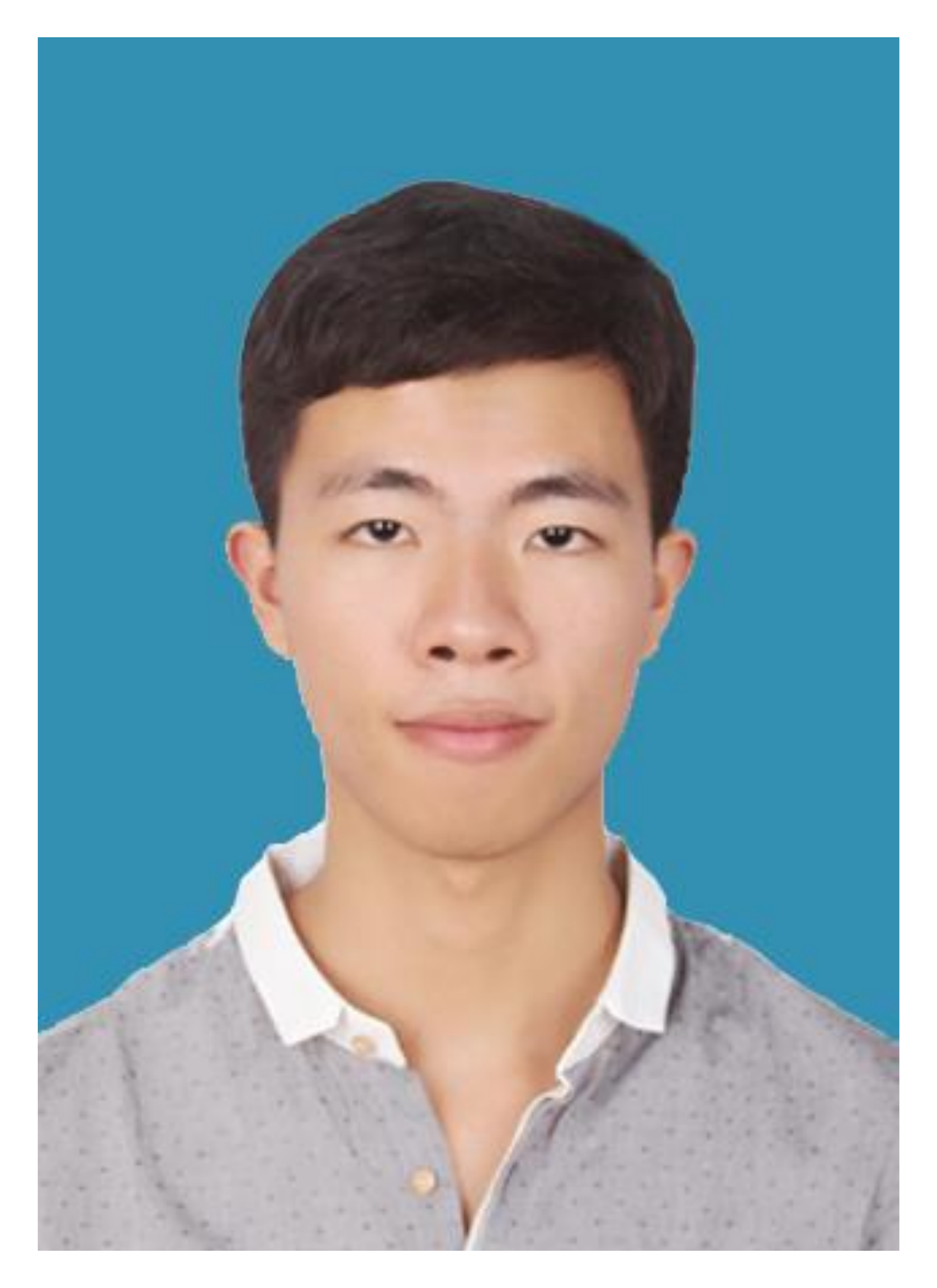}}]{Haipeng Zhang}
received the B.S. degree from the Department of Physics, Tsinghua University, Beijing, China, in 2016, and he is currently working toward the Ph.D. degree in the Department of Automation, Tsinghua University, Beijing, China.

His research interests include machine learning, deep learning and computer vision.
\end{IEEEbiography}

\vspace{-8mm}

\begin{IEEEbiography}[{\includegraphics[width=1in,height=1.25in,clip,keepaspectratio]{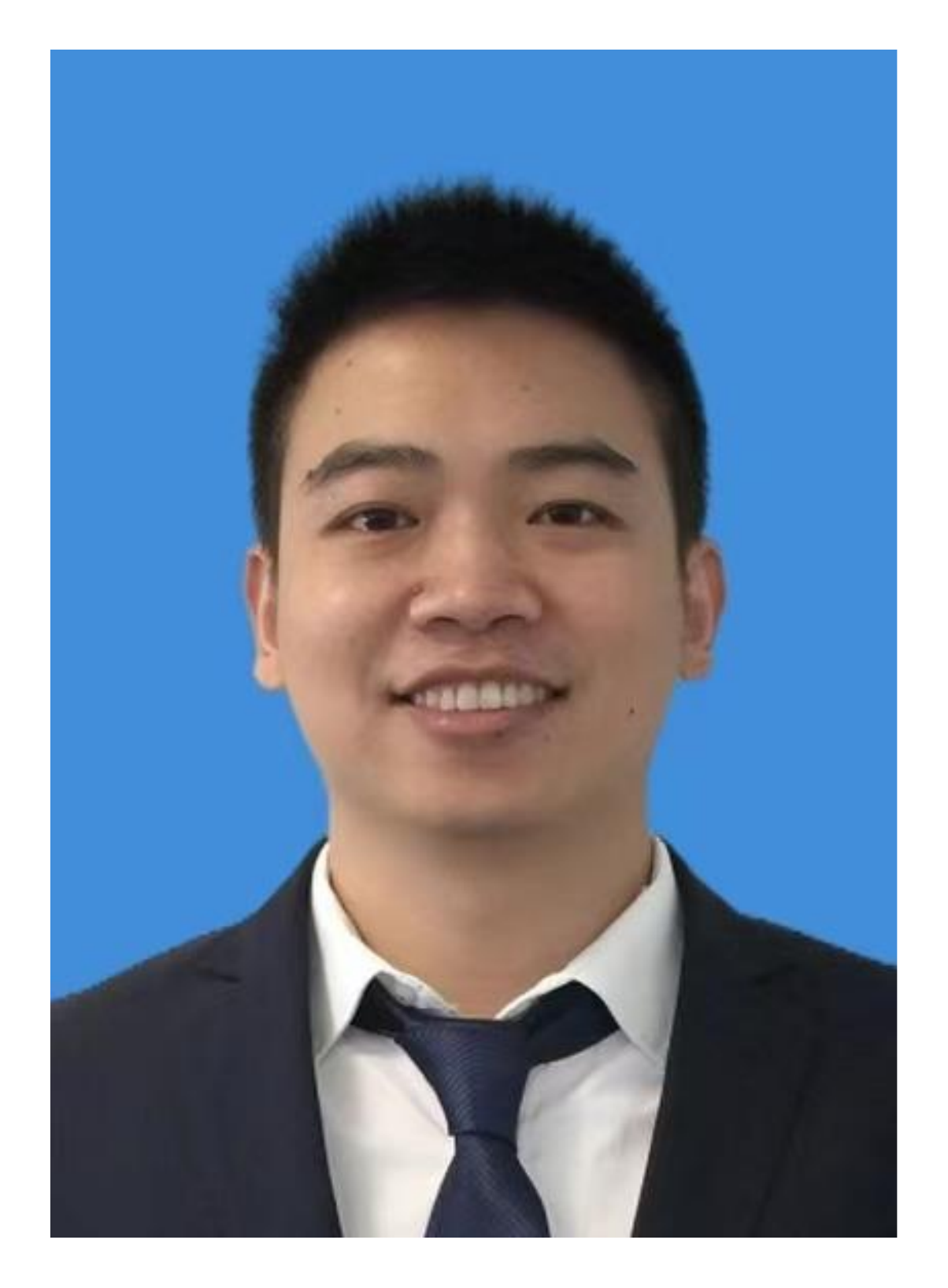}}]{Zhong Cao}
received the B.S. degree from the Department of Automation, Tsinghua University, Beijing, China, in 2014, and he is currently working toward the Ph.D. degree in the Department of Automation, Tsinghua University, Beijing, China.

His research interests include machine learning, deep learning and computer vision.
\end{IEEEbiography}

\vspace{-8mm}


\begin{IEEEbiography}[{\includegraphics[width=1in,height=1.25in,clip,keepaspectratio]{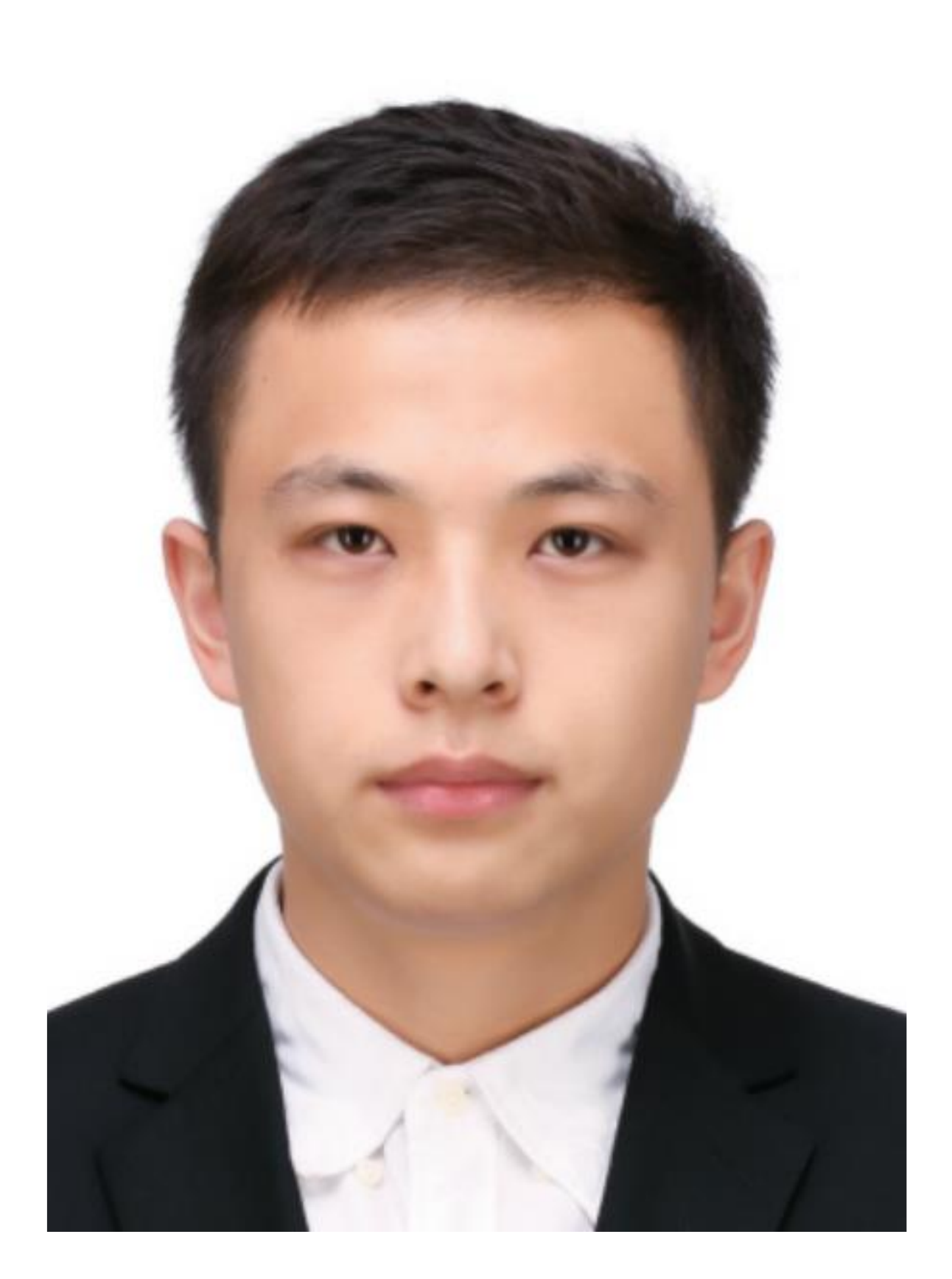}}]{Ziang Yan}
received the B.S. degree from the Department of Automation, Tsinghua University, Beijing, China, in 2015, and he is currently working toward the Ph.D. degree in the Department of Automation, Tsinghua University, Beijing, China.

His research interests include machine learning, deep learning and computer vision.
\end{IEEEbiography}

\vspace{-8mm}

\begin{IEEEbiography}[{\includegraphics[width=1in,height=1.25in,clip,keepaspectratio]{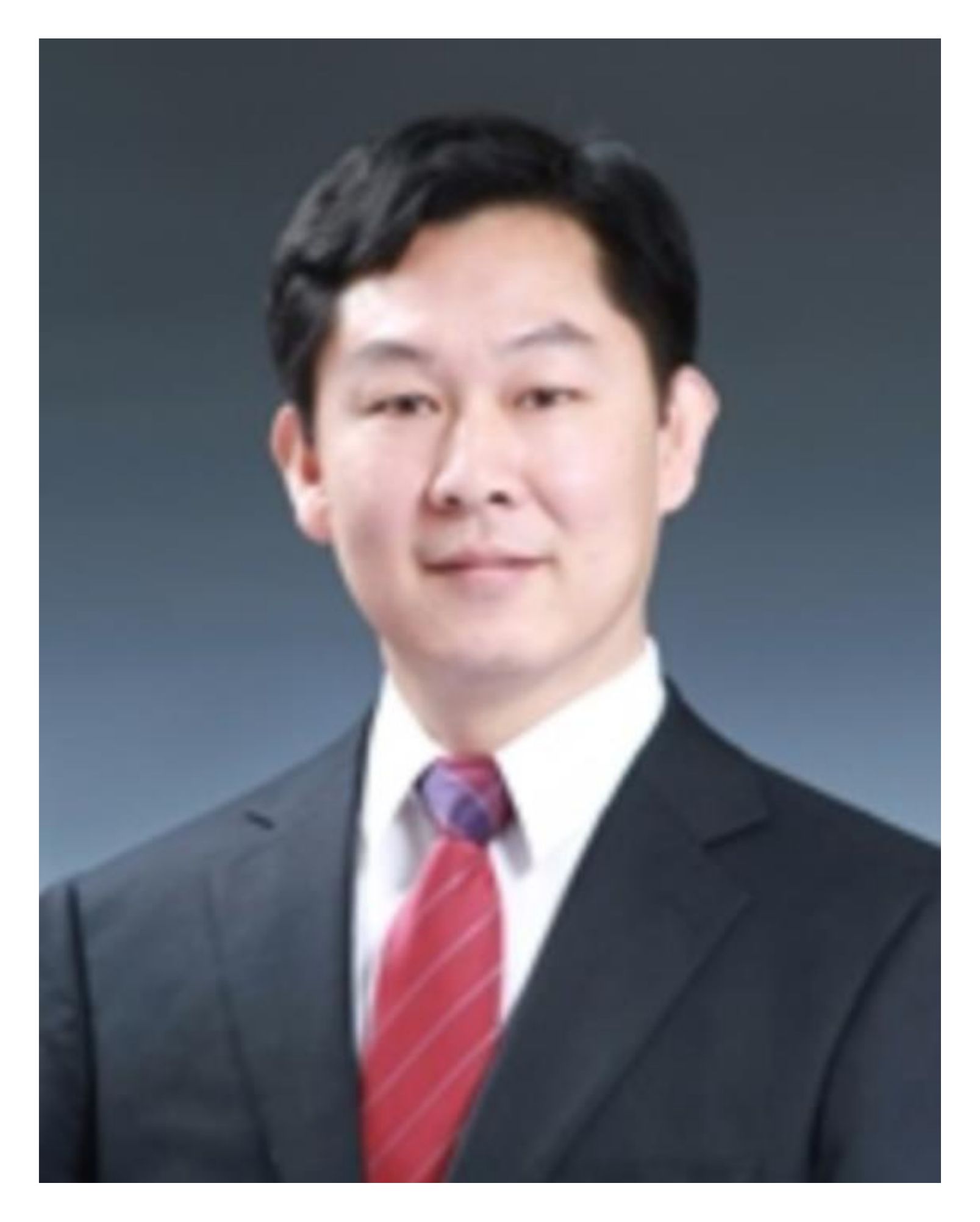}}]{Changshui Zhang}
(M’02–SM’15–F’18) received the B.S. degree in mathematics from Peking University, Beijing, China, in 1986, and the M.S. and Ph.D. degrees in control science and engineering from Tsinghua University, Beijing, in 1989 and 1992, respectively.

He is currently a Professor with the Department of Automation, Tsinghua University. His research interests include artificial intelligence, image processing and machine learning.
\end{IEEEbiography}




\end{document}